\documentclass[journal,twoside]{IEEEtran}
\usepackage{amsmath}
\usepackage{algorithmic}
\usepackage{array}
\usepackage{url}
\usepackage{color}
\usepackage[switch,columnwise]{lineno}
\usepackage{mathtools}
\usepackage[numbers,sort&compress,square]{natbib}
\newcommand{\etal}{\mbox{\emph{et al.\ }}}
\usepackage{amssymb}
\usepackage{todonotes}
\usepackage{graphicx}
\graphicspath{{images/}{../images/}}
\usepackage{subfigure}
\usepackage{xcolor}
\usepackage[export]{adjustbox}
\usepackage{relsize}
\usepackage{soul,xcolor}
\usepackage[hidelinks]{hyperref}

\usepackage[belowskip=-5pt,aboveskip=5pt,font=small]{caption}

\begin{document}
\setstcolor{red}

% \linenumbers

\title{Fast Enhanced CT Metal Artifact Reduction using Data Domain Deep Learning}

\author{Muhammad Usman Ghani, %~\IEEEmembership{Member,~IEEE,}
        W. Clem Karl,~\IEEEmembership{Fellow,~IEEE}
        \thanks{This material is based upon work supported by the U.S. Department of Homeland Security, Science and Technology Directorate, Office of University Programs, under Grant Award 2013-ST-061-ED0001. The views and conclusions contained in this document are those of the authors and should not be interpreted as necessarily representing the official policies, either expressed or implied, of the U.S. Department of Homeland Security.}
        
\thanks{M. U. Ghani and W. C. Karl are with the Department
of Electrical and Computer Engineering, Boston University, Boston,
MA, 02215 USA (e-mail: \texttt{\{mughani, wckarl\}@bu.edu)}.}

\thanks{This paper has supplementary downloadable material available at http://ieeexplore.ieee.org., provided by the author. The material includes additional experimental results and information. This material is 10.3 MB in size.}

}

\markboth{IEEE Transactions on Computational Imaging}%
{Ghani \MakeLowercase{\etal}: Fast Enhanced CT Metal Artifact Reduction using Data Domain Deep Learning}

\maketitle

\begin{abstract}
	Filtered back projection (FBP) is the most widely used method for image reconstruction in X-ray computed tomography (CT) scanners, and can produce excellent images in many cases. However, the presence of dense materials, such as metals, can strongly attenuate or even completely block X-rays, producing severe streaking artifacts in the FBP reconstruction. These metal artifacts can greatly limit subsequent object delineation and information extraction from the images, restricting their diagnostic value. This problem is particularly acute in the security domain, where there is great heterogeneity in the objects that can appear in a scene, highly accurate decisions must be made quickly, and processing time is highly constrained. The standard practical approaches to reducing metal artifacts in CT imagery are either simplistic non-adaptive interpolation-based projection data completion methods or direct image post-processing methods. These standard approaches have had limited success. Motivated primarily by security applications, we present a new deep-learning-based metal artifact reduction approach that tackles the problem in the projection data domain. We treat the projection data corresponding to dense, metal objects as missing data and train an adversarial deep network to complete the missing data directly in the projection domain. The subsequent complete projection data is then used with conventional FBP to reconstruct an image intended to be free of artifacts. This new approach results in an end-to-end metal artifact reduction algorithm that is computationally efficient textcolor{red}{and therefore} practical and fits well into existing CT workflows allowing easy adoption in existing scanners. Training deep networks can be challenging, and another contribution of our work is to demonstrate that training data generated using an accurate X-ray simulation can be used to successfully train the deep network, when combined with transfer learning using limited real data sets. We demonstrate the effectiveness and potential of our algorithm on simulated and real examples.
\end{abstract}


\section{Introduction} \label{intro}

\begin{figure}[tb]
	\centering
	\includegraphics[height=0.25\textheight]{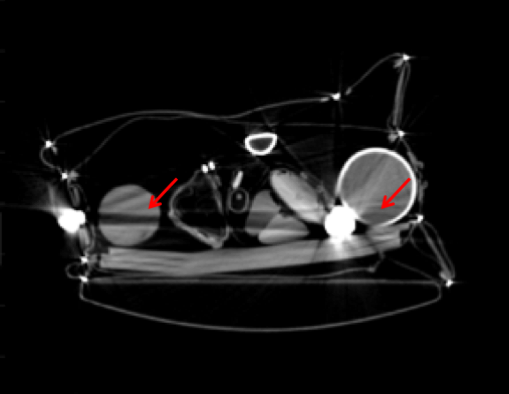}

	\caption{FBP reconstruction of a slice from a dataset collected to model airport checked-baggage scenarios is presented \cite{crawford2013research}. Severe metal artifacts have caused object splitting and boundary suppression which will result in inaccurate object segmentation and labeling.}
	\label{fig:intro}
\end{figure}

\IEEEPARstart{M}{etal} artifacts are common problems in conventional computed tomographic (CT) scanners using poly-energetic $X$-ray sources coupled with energy integrating detectors. When there is a dense metallic object in the field of view, it highly attenuates or completely blocks the $X$-rays for the corresponding detector locations. When these data are used by the conventional filtered back projection (FBP) method, the resulting image exhibits severe streaking artifacts and inaccurate CT numbers, as illustrated in Figure~\ref{fig:intro}. Such streaking then leads to errors in material or tissue type identification and object segmentations that can severely degrade automated threat recognition in security settings or clinical diagnosis in medical settings. Boas \etal \cite{boas2011evaluation} reported $21\%$ of medical scans in their study were subject to metal artifacts. The problem is worse in security applications, where metallic objects appear in many stream of commerce items \cite{martin2015learning,crawford2013research}. Thus, metal artifact reduction (MAR) is an important problem in CT imaging. \par

Motivated by the security application, this paper aims to reduce metal artifacts in CT images by applying deep learning (DL) directly in the projection-domain, prior to image formation. Unlike typical image post-processing approaches, by introducing learning in the projection data domain we aim to avoid the creation of artifacts altogether. Additionally, in our approach, metal-contaminated data is completely removed, therefore the type of metal is not critical. This approach also enables our method to use the conventional and efficient FBP approach for image reconstruction, which is fast and already implemented in most of the CT systems available today. We delete the metal-contaminated projection data and train a conditional generative adversarial network (CGAN) to perform projection data completion. The resulting ``enhanced" projection data is subsequently used with the FBP method to reconstruct the CT image. This new deep-learning-based MAR (\emph{Deep}-MAR) approach results in an end-to-end framework that is computationally efficient and tractable, and which fits well into existing CT workflows, allowing easy adoption in existing scanners.  In this initial work, we focus on $2$-dimensional (planar) problems. A preliminary version of our learning-based sinogram completion approach was presented in \cite{ghani2018deep} for sparse-view tomography.

\subsection{Contributions}
\subsubsection{\emph{Deep}-MAR framework}
The main contribution in this work is the new \emph{Deep}-MAR framework that uses state of the art deep learning (DL) methods based on convolutional neural networks (CNNs) to complete projection data in a tractable and computationally efficient way. 

\subsubsection{CGAN-based sinogram completion} The second contribution of this work is the application of state-of-the-art conditional generalized adversarial networks (CGANs) for sinogram completion. CGANs have shown incredible power and generalizability in a variety of image processing problems and yet are relatively new to inverse problems in general and CT in particular. In order to make our work reproducible, simulated test examples, test codes, and trained models have been publicly made available\footnote{\url{https://codeocean.com/capsule/0510438}}.\par

\subsubsection{Network training with simulated data and transfer learning} The availability of adequate training data is a fundamental DL challenge. This challenge is especially acute for security related CT data, where the universe of possible scenes is very large and thus limits what can be obtained through direct physical experiment. The third contribution of this work is demonstrating the use of a large set of simulated data based on accurate $X$-ray physics coupled with transfer learning exploiting a more limited amount of real experimental data.  We have made the simulation setup and simulated data prepared for this work publicly available\footnote{\url{https://github.com/mughanibu/DeepMAR/}}. This will motivate and help researchers to develop new machine-learning-based methods to improve CT image reconstruction and analysis.\par

\subsubsection{Pathway to generating real training datasets} The fourth contribution of this work is the ability to generate real projection-domain training data pairs matched to our framework. By casting the learning problem as one of sinogram completion rather than direct correction we focus the deep network on the task of missing data generation. This focus means the network does not need to handle all the different possible types of metal that can appear in a sinogram, just the lower-dimensional geometric missing-data configurations. As a consequence, our network only needs matched pairs of sinograms with missing and corresponding complete (but non-metal) data for training. This structure makes it possible to generate additional real-data physical examples to train the DL-based projection-domain MAR method from existing scene data without metal objects through simple sinogram data deletion, a much easier task.\par

\subsection{Prior-Work}

Existing MAR algorithms can be  grouped into  three  categories: Model-based iterative reconstruction (MBIR-MAR), image-domain MAR (ID-MAR), and projection-domain data completion (PD-MAR). MBIR-MAR methods incorporate a physically derived observation model together with appropriate image-domain priors in an iteratively-solved optimization problem \cite{elbakri2002statistical,wang1996iterative,de2000reduction, hamelin2008iterative, jin2015model,crawford2013research,chang2018prior}. MBIR methods can produce excellent results and are a principled way of incorporating prior information. Unfortunately, they are typically computationally expensive, requiring repeated reconstruction and forward projection in an iterative algorithm. This computational cost has unfortunately limited their practical impact to date, especially in security applications where throughput is important. \par

Image-domain MAR (ID-MAR) methods rely on image-processing techniques as a post-processing strategy to reduce streaks in the reconstructions. ID-MAR methods first reconstruct images which contain unwanted artifacts, and then attempt to correct these artifact-filled images. The problem with this strategy is that once important image structure is lost to artifacts, it can be very difficult to recover it effectively. In the medical context, Soltanian-Zadeh \etal \cite{soltanian1996ct} estimate streaks by subtracting a low-pass-filtered version of an image from itself and then thresholding the difference image. Streaks are removed by subtracting this estimate from the original image. This technique involves an ad-hoc thresholding step and does not attempt to recover structure lost in the streak areas. In the security context, Karimi \etal \cite{karimi2015metal} estimated streaks by computing the difference of penalized least squares (PLS) and penalized weighted least squares (PWLS) reconstructions where metal-contaminated projections are down-weighted and the estimated streaks are later removed from the original reconstruction using subtraction. In the medical context, Gjesteby \etal \cite{gjesteby2017deep} recently proposed coupling conventional normalized MAR (NMAR) with a convolutional neural network. Their method attempts to learn a mapping from the NMAR image to an artifact-free image, with the aim of eliminating the residual artifacts of the NMAR method. Training was done on phantom images with artificially inserted metal objects. The data used appeared limited to $40$ slices of a single case, with $5$ slices held out for testing. Extending the approach to incorporate more general real data would appear challenging, since obtaining matched image pairs with and without various metal materials would be difficult. Since the nature of the artifacts will change with the nature of the metal in the scene, it would seem that the training set would need to cover the range of metal materials anticipated in addition to the configurational variability, which would be difficult to achieve in the security context.\par

PD-MAR approaches aim to replace the metal-corrupted projection data with data obtained using neighboring information. These methods basically treat metal-corrupted projection data as missing data and attempt to complete these projections. Kalender \etal \cite{kalender1987reduction} proposed a simple approach to replace the metal-corrupted data using the one-dimensional linear interpolation (LI) of neighboring detector channels. This approach is by far the most common MAR scheme and is considered a benchmark \cite{gjesteby2016metal}. However, its performance deteriorates as the size and number of metal objects increases. In attempting to ameliorate the original metal streaking it can also introduce new streaking artifacts, as can be seen in the results in Section~\ref{res}. In an attempt to improve on the LI method, Mahnken \etal \cite{mahnken2003new} proposed an extended, two-dimensional interpolation method that replaces the metal-corrupted projection data with a weighted sum of $N$-nearest neighbor data points. Rivi{\`e}re \etal \cite{la2006penalized} proposed a statistical objective function-based projection data restoration method that iteratively corrects for data and model mismatch. The method is not focused on data completion or metal artifacts and can be computationally expensive.  

NMAR \cite{meyer2010normalized} is yet another PD-MAR algorithm that relies on the presence of a prior image similar to the one of interest and combines this prior with a simple interpolation scheme. Zhang \etal \cite{zhang2017convolutional} recently proposed a similar approach, in which a prior image is estimated using a trained CNN. Gjestsby \etal \cite{gjesteby2017deepNMAR} trained a CNN to learn mapping from NMAR sinogram to metal-free sinogram. These approaches rely heavily on the prior image and their performance degrades if an appropriately matched prior image is not available. Such prior image-based correction methods are particularly problematic for security scenarios because a good prior image is typically not available due to the high variability of objects and materials present in such scenes. \par

Claus \etal \cite{claus2017deep} recently proposed a learning-based PD-MAR approach. They trained a fully connected neural network to perform interpolation of missing data in the sinogram domain, and is thus similar in focus to the current work. Their method applies to a highly constrained problem with only a single metallic object of fixed and known size placed at the very center of the field of view. Since the network is trained for this very constrained situation, it is not clear how this method could be extended to realistic scenarios with multiple metal objects placed at arbitrary locations. In addition, its reliance on a fully connected network severely limits scaling to practical problem sizes. Overall, while there has been initial work aimed at applying learning in the CT data domain, these existing works have been limited to small, toy examples and/or highly constrained scenarios not representative of real applications.\par

Anirudh \etal \cite{anirudh2017lose} proposed an implicit sinogram completion method for limited angle CT application. In contrast to our approach, their method consists of a network that learns a mapping from incomplete projection data to a final complete-data image. They demonstrated that their approach produces better reconstructions and segmentations using incomplete data as compared to conventional limited angle CT methods. 

\begin{figure*}[tb]
\centering
\includegraphics[width = 0.99\textwidth]{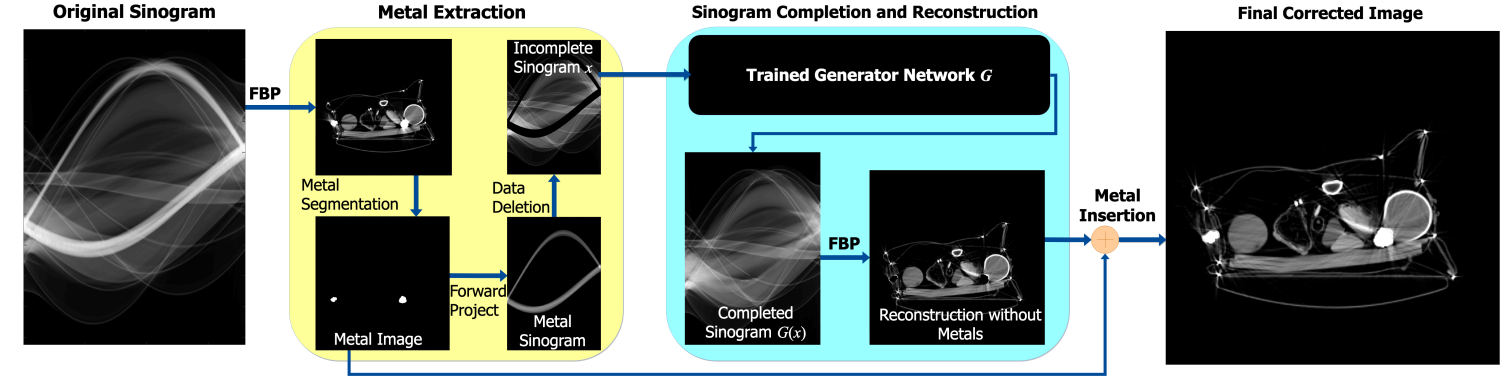}
\caption{Work flow and major components of our \emph{Deep}-MAR framework are presented. We start with a metal-corrupted sinogram which is used to create an FBP reconstruction to identify the metal objects in the image and delete the corresponding metal-contaminated data in the sinogram. Incomplete sinogram is input to the generator network to complete the sinogram which is later used for FBP reconstruction.}
\label{fig:deep_mar}
\end{figure*}

Learning methods based on CNNs have had impact in a variety of applications. CNNs have demonstrated impressive performance on various image restoration tasks, including image super-resolution \cite{kim2016accurate}, image denoising \cite{zhang2017beyond}, and artifact reduction \cite{jin2017deep}. However, training CNNs with a mean-square-error loss function has been shown to result in over-smoothed images \cite{pix2pix17,zhao2017loss}. Adversarial training is an elegant strategy to train neural networks \cite{goodfellow2014generative} and Isola \etal \cite{pix2pix17} suggest that a traditional loss coupled with an adversarial loss is a better overall loss function for \emph{image-to-image translation} tasks. They proposed an approach called a Conditional Generalized Adversarial Network (CGAN) for these tasks. A CGAN consists of two networks: a generator network that learns to perform \emph{image-to-image translation}, and a discriminator network that is trained to discriminate artificial from true images. The discriminator network is used as an adversarial loss for training the generator network, which forces the generator to be better and better at its job in order to fool the discriminator. We use generator and discriminator network architectures that are inspired from Isola \etal \cite{pix2pix17}. Different CNNs have already been successfully applied to perform missing data replacement or completion in the image domain \cite{pix2pix17, kohler2014mask,chaudhury2017can,xie2012image}. However, using CNNs for sinogram completion amounts to learning \emph{sinogram-to-sinogram translation}, which is a different task. \par

The success of CNNs depends upon successful learning of internal representations, which typically requires large datasets. This need poses a major challenge for the application of CNNs to many areas where access to large datasets is not possible due to privacy or security concerns or the need for highly skilled labor to provide ground truth annotations, e.g., medical imaging, and especially security applications. A recent trend in these areas is to use transfer learning \cite{yosinski2014transferable}. In this work, we investigate the use of transfer learning in the security context by pre-training the deep network using a large \emph{simulated} dataset and then fine-tuning the network on a modestly sized real dataset.\par

\section{Learning-Based MAR Approach}  \label{meth}

In this paper, we propose a framework we term \emph{Deep}-MAR to reduce metal artifacts in CT images using adversarial deep learning to perform completion of missing projection data in the sinogram domain (i.e. sinogram completion). The \emph{Deep}-MAR framework is motivated by and focused on reducing the effects of metal in checkpoint security imagery. 

\subsection{Deep-MAR Algorithm}
The major steps of our \emph{Deep}-MAR framework are presented in Figure~\ref{fig:deep_mar}. There are three major steps: i) identification and suppression of metal-contaminated projection data, ii) CGAN based sinogram completion, and iii) efficient FBP image reconstruction and reinsertion of metal objects back into the reconstructed image. 

In order to identify the metal-contaminated projection data a conventional FBP reconstruction using the original sinogram is first generated. Thresholding followed by morphological operations are then used to segment the metallic objects in the image, which are forward projected to generate a mask, $M$, in the sinogram domain corresponding to the metal traces. This mask is made slightly larger than the original thresholding to be sure to remove all the metal. The metal-contaminated projection data is then deleted from the original sinogram. Similar metal segmentation methods have been used in many of the MAR approaches \cite{mahnken2003new, meyer2010normalized, do2014sinogram, karimi2015metal}. One could potentially use more sophisticated metal segmentation approaches, for instance a DL-based approach for object segmentation \cite{long2015fully}.

\begin{figure*}[tb]
\centering
\includegraphics[width = 0.99\textwidth]{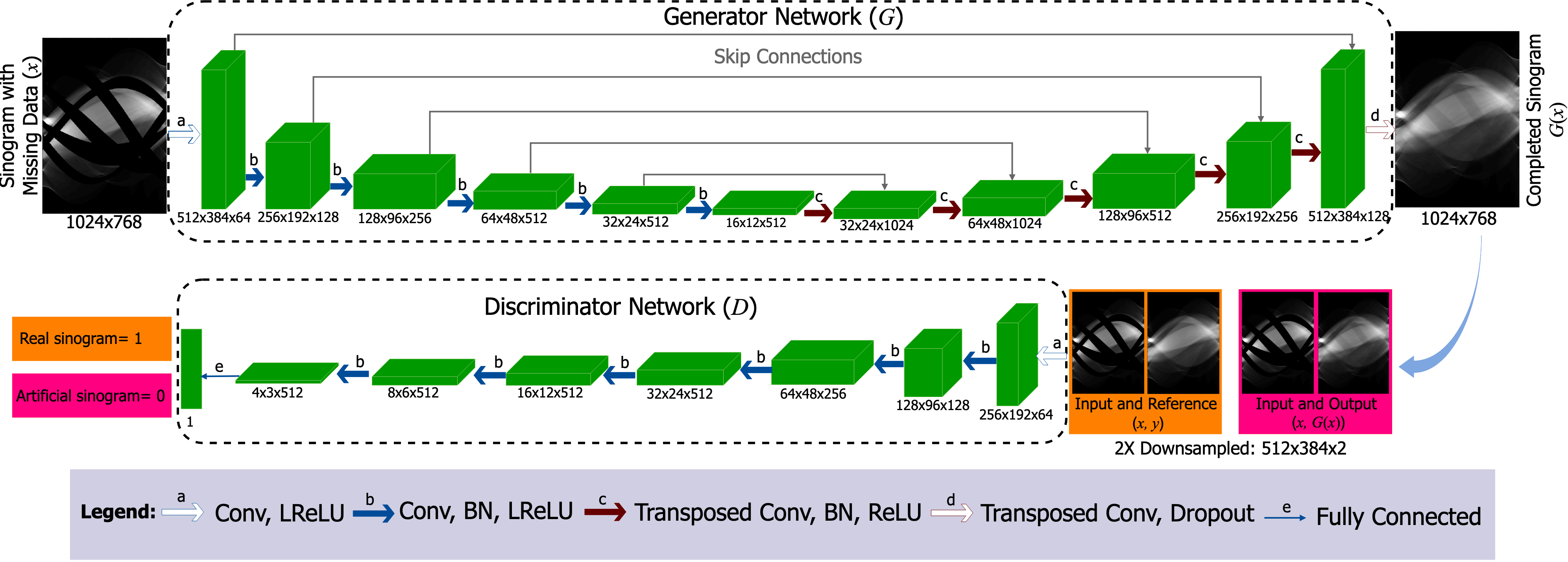}
\caption{Overall CGAN framework consisting of a generator and a discriminator network is presented. The generator network follows a $U$-Net-like architecture, where the input is an incomplete sinogram and it learns to complete the sinogram. The discriminator network learns to distinguish between ground truth and generator completed sinograms, and is used as an adversarial loss term to train the generator network so it can produce realistic output. Abbreviated legend at the bottom of the Figure are defined here, Conv: $2$D convolutions, Transposed Conv: transposed convolutions, BN: batch-Normalization, LReLU: leaky rectified linear units, ReLU: rectified linear units.}
\label{fig:cgan}
\end{figure*}

\subsection{Learning-Based Sinogram Projection Completion}

The heart of the \emph{Deep}-MAR framework is a state of the art CGAN, which is used to perform the sinogram completion task. A CGAN is composed of two distinct networks: a generator network $G$ and a discriminator network $D$. The two networks are trained by optimizing an objective function based on a mini-max game \cite{pix2pix17,goodfellow2014generative}. The overall mini-max objective function we use is given by:
\begin{equation}
	G^* = \arg \min\limits_G \max\limits_D \mathcal{L}_{cGAN}(G,D) + \lambda \mathbb{E}_{x,y} [\| y - G(x) \|_2^2]
	\label{eq:overall}
\end{equation}
where, $x$ and $y$ represent input and ground truth pairs of incomplete-data and true complete-data sinograms, respectively; $G(x)$ is the estimated complete-data sinogram produced by the generator; $\mathbb{E}_{x,y}$ defines expectation over data density $p_{data}$ approximated by training data pairs $x$ and $y$, $D$ denotes the discriminator network, and $\lambda$ is a hyper-parameter used to control the balance between the two terms in the expression. The inclusion of a discriminator loss was seen to improve the overall result relative to an $\ell_2$ only loss. In the Supplementary Material, we provide more detail on the value of the $\ell_2$ and discriminator loss terms to the overall formulation.

In the optimization (\ref{eq:overall}) the networks $G$ and $D$ act as adversaries -- that is, $G$ attempts to make perfect sinograms while $D$ attempts to detect fakes. The first term in the objective captures this interaction through an adversarial loss between $G$ and $D$. This loss is given by:
\begin{equation}
	\small
	\mathcal{L}_{cGAN}(G,D) = \mathbb{E}_{x,y} [\log D(x, y)] +  \mathbb{E}_{x} [\log (1 - D(x, G(x))]
\end{equation}
were, $\mathbb{E}_{x}$ defines expectation over incomplete data density, approximated by averaging over input training data samples $x$. The presence of this loss forces the $D$ network to improve its discrimination ability and thus also forces the generator network $G$ to become better and better at completing sinograms. In other image processing tasks, such an adversarial approach has been shown to produce more robust and higher performance generator networks $G$. The second term in the optimization (\ref{eq:overall}) is the traditional loss associated with fit to the training data. In this case we use the $\ell_2$ loss because our initial experiments indicated that an $\ell_2$ loss performs better for sinogram completion, as compared to choices such as the $\ell_1$ loss. 

The internal structure of the coupled generator and discriminator network architectures are presented in Figure~\ref{fig:cgan}. Convolutional kernels of size $5 \times 5$ with a stride of $2$ are used at each layer for both the generator and discriminator networks. The generator network, $G$, has a modified U-Net architecture with a fully convolutional architecture. Instead of using max-pooling layers for down-sampling, two-pixel strided convolutions are used. For up-sampling, transposed convolutions with two-pixel strides are used. Overall $6$ down-sampling and $6$ up-sampling layers were chosen. Additionally, there are skip connections between each layer $i$ and $n-i$, where $n$ is the total number of layers. The skip connection at each layer merely concatenates outputs at layer $i$ to layer $n-i$. Using stride-2 convolutions result in sub-sampling and significantly larger practical effective receptive field (ERF) as compared to stride-1 convolutions \cite{luo2016understanding}. The theoretical ERF grows with the number of layers -- $6$ down-sampling and $6$ up-sampling layers result in a theoretical ERF of $757\times757$ pixels \cite{kayalibay2017cnn}, so each estimate could potentially use information from very large area of the input sinogram. Kernel size selection is also guided by the theoretical ERF in that using smaller kernels would require more layers to achieve a desired ERF and larger kernels would result in many more learning parameters needing estimation. Empirical testing showed that $5\times5$ convolutional kernels with a shallower architecture perform better than $3\times3$ convolutional kernels with a deeper architecture. Dropout is used in the last layer to avoid over-fitting. It is used after all batch-normalization (BN) layers as suggested by Li \etal \cite{li2018understanding} since dropout results in variance shift when applied before BN. \par

The discriminator, $D$, is based on the full sinogram image, instead of just patches. Given a ground truth or network output sinogram conditioned on the input incomplete sinogram, our discriminator network classifies the full sinogram as real or fake. The reason behind using a full-sinogram discriminator is that missing data follows sinusoidal structure, which would not be possible to accurately capture using a patch-based discriminator. Additionally, the network is trained to perform non-blind sinogram completion. The missing data are the metal-contaminated data points in sinogram, which are masked before the sinogram is input to the generator network. By using mask-specific sinogram completion our loss function is focused on the areas of most interest -- where the data is missing. The generator network output is given by: $G(x) = x + M \odot  x_{\texttt{D1}}$, where, $x_{\texttt{D1}}$ is the output of the last layer of the generator network, $M$ is the metal mask, and $\odot$ denotes element-wise multiplication. This choice retains non-metal sinogram data as is, and uses the deep network to replace only the metal traces. We use $2X$ downsampled sinograms for discriminator ($D$) to speedup the training procedure.  \par

\section{Network Training} \label{sec:training}

A major challenge in the use of deep network architectures is having sufficient data to allow robust training of the network. In many computer vision and image processing tasks there is access to abundant samples of training data (e.g. ImageNet \cite{deng2009imagenet}, which contains over $14$ million images). In contrast, in the security domain there is a great diversity of possible objects in a scene, yet the amount of physical CT data available for training is severely limited. To address this limitation we generated a training set for sinogram completion using physically accurate $X$-ray simulation tools. After initial training, our networks are then refined with available physical data through transfer learning. Areas without metal are considered background (air) in this work -- that is, the target replacement value for metal regions is background material and, conversely, only locations with background material are considered for virtually embedding metals in real-data. This approach makes the problem more tractable since the network can focus on a smaller space of possibilities. 

\subsection{Simulated Training Data Generation} 
Most conventional CT scanners use a poly-energetic $X$-ray source coupled with energy integrating detectors. We can accurately model the data obtained from such scanners as a sum of mono-energetic sinograms weighted by the corresponding relative strength of the source spectrum. The resulting projection output can thus be well modeled as:
\begin{equation}\label{eq:poly_sino}
\begin{aligned}
    &\Tilde{I}^{(i)} = I_0 \cdot \eta \Big(E^{(i)}\Big) \\
    &I_j^{(i)} = \Tilde{I}^{(i)} \cdot  e^{- \int_{L_j}  \mu( \vec{x}, E^{(i)})\: d\ell}
\end{aligned}
\end{equation}
where $\Tilde{I}^{(i)}$ is the relative $X$-ray source intensity at energy $E^{(i)}$, obtained from the energy-dependent scanner source weighting function $\eta (E^{(i)})$ at energy $E_i$ and scaled by the overall $X$-ray source intensity or blank scan factor, $I_0$. The sinogram contribution at energy $E^{(i)}$ for ray-path $j$ corresponding to the line $L_j$ is denoted by $I_j^{(i)}$ and is obtained from the Beer-Lambert law using the energy-dependent scene attenuation coefficient $\mu(\vec{x}, E^{(i)})$ at location $\vec{x}$ and energy $E^{(i)}$ integrated over the associated ray path. Note that the exponent in the Beer-Lambert expression is a line integral projection, and is approximated by discretizing the integral using standard methods. In this work we use a common ray-based projector model \cite{jin2015model}, though other models are certainly possible (e.g. distance driven projection, Fourier methods, etc.).

The projection data contributions are degraded by data-dependent and electronic noise. The data-dependent variability follows a Poisson distribution with mean corresponding to $I_j^{(i)}$. The electronic noise is modeled as Gaussian zero-mean and variance $\sigma_e^2$. Since the standard CT data are log-normalized, the final $X$-ray observation model for ray-path $j$ is defined as:
\begin{equation}\label{eq:poly_norm}
	y_j 
	= - \ln \left(\frac{ \sum\limits_{i=1}^{N} \left( 
	g^{(i)} \text{Poisson} \left( I_j^{(i)} \right) \right)
	+ 
	\mathcal{N}(\underline{0}, \underline{\sigma_e^2})}{
	\sum\limits_{i=1}^{N} g^{(i)}\Tilde{I}^{(i)} 
	}\right)
\end{equation}
where the sum is over the contributions from each of the $N$ source energies used to approximate the continuous spectrum. The term $g^{(i)}$ is the energy-dependent detector response or gain which captures the photon to charge conversion factor, therefore, $y_j$ is measured in units of current. This model is consistent with poly-energetic models used in different published studies \cite{crawford2013research,elbakri2002statistical,la2006penalized,chang2017modeling}. By incorporating a full energy aware ray model, effects such as beam hardening are inherently included. Note that one could consider other sensing models and configurations in a straightforward way by instantiating appropriate simulation models, including, for example, photon-counting detectors, Monte-Carlo-based photon scatter, etc.

\begin{figure}[tb]
\centering
\includegraphics[width=0.5\textwidth]{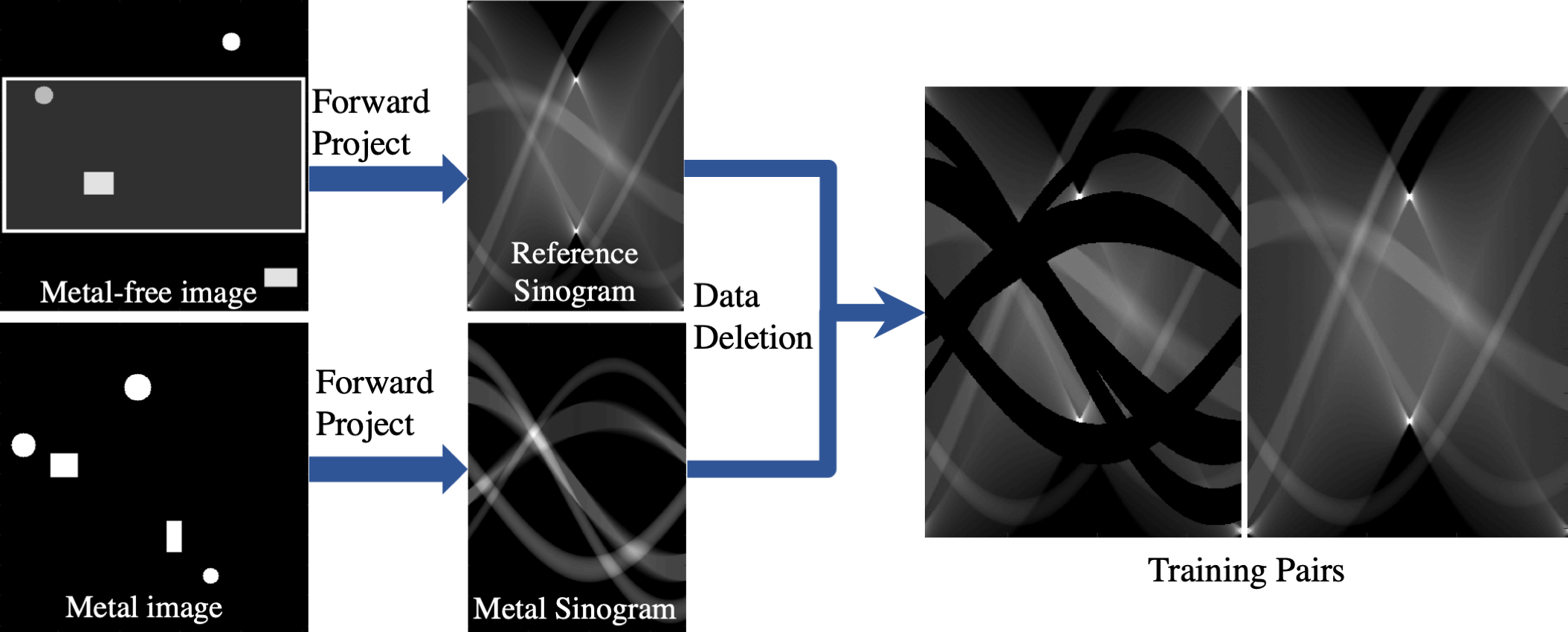}
\caption{Setup for simulated training pair generation. A metal-free scene is created using our stochastic bag simulator which is forward projected to compute a reference sinogram using our physically accurate $X$-ray scanner model. Our stochastic bag simulator then defines geometrical configurations for metallic objects in the scene to generate masks for data deletion in the sinogram domain. These masks are then used to delete data resulting in an incomplete sinogram. A large set of training pairs consisting of incomplete and reference sinograms are created this way to train our CGAN model.}
\label{fig:train_data_gen}
\end{figure}

Coupled with this physically accurate $X$-ray scanner model, we also created a stochastic bag simulator. This bag or scene simulator places objects of different material composition and of varying geometric configuration at random locations in the scene. The current instantiation of the simulator uses geometric primitives including circles and rectangles and uses material attenuation coefficients drawn from the NIST XCOM database \cite{berger1987xcom}. The setup for simulated training dataset generation is illustrated in Figure~\ref{fig:train_data_gen}. A metal-free scene is created and then multiple metallic object areas are defined with corresponding sinogram data deletions to create multiple training pairs. These training pairs consist of sinograms with data deletions corresponding to metal object locations (incomplete sinograms) and the corresponding sinograms without the metal objects (reference completed sinograms). The entire process can be done quickly allowing the generation of tens of thousands of sinogram pairs, much more than would be possible to do physically. Other, more complex, scene generators could also be used for this purpose. 

\subsection{Real Training Data Generation} \label{sec:real_data_train}

Acquiring physical matched sinogram pairs with and without metal objects for training is a challenge, especially in security settings. What would be required in general is a real sinogram containing metal objects and the corresponding real sinogram if those objects were absent, but everything else in the scene remained the same. Such consistency is very difficult to achieve in practice. 

As a consequence of formulating the PD-MAR problem as one of sinogram completion, our approach only requires metal-trace masked and corresponding metal-free complete sinograms for CGAN training since we are focusing our learning on missing data completion. We can use sinogram data from scenes which do not contain metal objects, and then create matched sinograms by deleting data in the sinogram that corresponds to geometric configurations of embedded objects. We define these object configurations at locations of the image associated with voids or background. In this way we can manually generate a modest a number of real pairs of incomplete and complete sinograms to train our CGAN. The security scans used for this data generation were obtained from the ALERT Center of Excellence (http://www.northeastern.edu/alert/transitioning-technology/alert-datasets/)

\subsection{Real Data Transfer Learning}

We combine our two sources of training data (that generated by accurate $X$-ray physics simulation and that derived from real scans) through transfer learning from the larger simulated data set. Transfer learning involves transfer of knowledge from a learned function, $f_S(\cdot)$, that solves a task $\mathcal{T}_S$ in a source domain $\mathcal{D}_S$ to achieve improved performance in learning a function, $f_T(\cdot)$, to perform a task $\mathcal{T}_T$ in a target domain $\mathcal{D}_T$ \cite{pan2010survey,yosinski2014transferable}. The most common use of transfer learning in DL research involves learning a function, $f_S(\cdot)$, in the form of a deep network from a large set of training data and then copying the first $k$ layers of this network $f_S(\cdot)$ to a target network $f_T(\cdot)$ as an initialization. In general, $\mathcal{D}_S \neq \mathcal{D}_T$ and the target task $\mathcal{T}_T$ may or may not be the same as the source task $\mathcal{T}_S$ \cite{yosinski2014transferable}. The initial network then undergoes additional refinement or training using a smaller set of training data from the new task. Transfer learning has been shown to provide improved performance, generalization, and robustness to over-fitting as compared to random initializations combined with limited training for various image classification tasks even when $\mathcal{D}_S$ and $\mathcal{D}_T$ are very different from each other \cite{yosinski2014transferable,gutman2016skin,rodner2015fine,malmgren2017improving}. \par

In this work transfer learning from a larger set of simulated data to a smaller set of real data is used. The learning task (sinogram completion) remains same in our case, i.e., $\mathcal{T}_S=\mathcal{T}_T$, however, the nature of the training data is changed, i.e., $\mathcal{D}_S \neq \mathcal{D}_T$. Additionally, we use the same network architecture for both source and target tasks, and we copy all layers of $f_S(\cdot)$ to $f_T(\cdot)$. We first train the network on a large simulated dataset acquired by the physically accurate simulation setup previously described. This trained network, $f_S(\cdot)$, is later fine-tuned using a smaller amount of real data. This strategy allows us to achieve good performance on real data even with a small real dataset.\par 

\section{Experiments} \label{exp}
This section describes the details of our experiments, networks, and training. For simplicity, we restrict our experiments to 2-D parallel-beam geometry in this work. Training deep networks is a challenging and time-intensive task, but once trained, inference is very fast. Note that in the examples considered here our proposed \emph{Deep}-MAR method is very efficient; our generator network $G$ takes approximately $58$ milliseconds to perform sinogram completion on an NVIDIA Tesla P100 GPU at test time, which makes it attractive in real world contexts. We use a CPU implementation of FBP which has not been optimized for efficiency, and on average it takes $1.55$ seconds using an Intel Xeon ($2.4$ GHz) processor on a Linux system with MATLAB R2017b to reconstruct the 2-D images considered in this work. \par

\subsection{Metal Artifact Dataset}
\subsubsection{Simulated Dataset}
Our simulated data model is based on the Imatron C$300$ scanner. Equation \ref{eq:poly_norm} was used to simulate poly-energetic CT with a weighting function $\eta (E^{(i)})$ and detector response $g^{(i)} = 2.6\times 10^{-3} \cdot E^{(i)}$ pA/quanta estimated by Crawford \etal \cite{crawford2013research}, source intensity set to $I_0 = 1.7 \times 10^5$ photons per ray, and using $N=121$ uniformly sampled energy levels between $10$ KeV and $130$ KeV. The electronic noise variance $\sigma_e^2=3.37 \text{pA}^2$ is modeled as counting statistics of $20$ photons detected at $65$ KeV following \cite{crawford2013research}. Photon scatter could certainly be included in our simulation setup in a straight forward way. However, accurate Monte Carlo scatter simulation methods are computationally very expensive ($O(n^6)$) \cite{ruhrnschopf2011general}. This additional computational load would limit the amount of simulated training data we could generate. We decided a simpler ray-based simulation was sufficient for sinogram completion training that was required and therefore, in this initial work, we do not include scatter in our simulation setup. Attenuation images of size $475mm \times 475mm$ were generated with attenuation coefficients from the NIST XCOM database \cite{berger1987xcom}. The list of metallic and non-metallic materials used is presented in the Supplementary Material (Section S1). Sinograms at each of the $N=121$ energy levels were generated using $720$ uniformly sampled projection angles between $0^0$ and $180^0$, and $1024$ detector channels per projection angle. In order to match the network input and output size, zero-padding was performed to result in a sinogram of dimensions $1024\times768$. We generated $10{,}000$ example images using the stochastic bag simulator with non-metallic objects, and placed up to $5$ metallic objects in each instance resulting in a training dataset of $50{,}000$ sinograms. An additional $25$ suitcase scenes with different configurations of cylinder and sheet objects (similar to the ones in the ALERT Task Order $3$ dataset \cite{crawford2013research}) were manually generated and included in the training dataset. For each of these images, matched pairs of sinograms without and with metallic objects in the scene were generated. We used horizontal flips of this data as a data augmentation strategy. The ASTRA toolbox \cite{van2016fast} was used for accelerated forward projection in the training data preparation. We used our stochastic bag simulator to generate $102$ reference metal-free scenes and inserted up to $5$ metals to result in $510$ test dataset sinogram pairs which were used for qualitative and quantitative analysis of our approach on simulated dataset. \par

\subsubsection{Real Training Dataset} 
Additional real training data was generated from real sinogram data acquired by an Imatron $C300$ scanner. The data was part of a collection effort supported by DHS \cite{crawford2014advances}. The scans were performed with a field of view (FOV) of $475mm \times 475mm$ with a peak $X$-ray source energy level of $130$ KeV. Data was re-binned to a parallel geometry with $720$ projection angles and $1024$ detector channels. Further, to match the network input and output size, zero-padding was done to result in a sinogram of dimensions $1024\times768$. In order to create the pairs of sinograms for training of the CGAN, $1{,}706$ different slices which did not contain significant metal objects were manually identified from the scanned bags. The strategy described in Section~\ref{sec:real_data_train} was used to generate $8{,}530$ pairs of real data sinograms for training with sinogram missing data corresponding to up to $5$ virtual metallic objects. Horizontal flips of this data were also used as a data augmentation strategy. This real data was used for transfer learning as previously described. \par

\subsubsection{Real Testing Dataset} \label{real_testing}
For quantitative analysis of performance on real data, $159$ pairs of sinograms were generated by selecting metal-free images from the ALERT Task Order $4$ dataset \cite{crawford2014advances} using the strategy described in Section~\ref{sec:real_data_train}. This strategy was used for quantitative analysis since the reference artifact-free images without any metal objects are available in this settings. For qualitative analysis on real data, slices were selected from the ALERT Task Order $3$ dataset \cite{crawford2013research} which contained metals and caused severe artifacts in the reconstructions. FBP reconstructions were generated for selected slices, and a metal mask identified by applying a threshold of $4000$ Modified Hounsfield Units (MHU $= 1000 + 1000(\mu - \mu_{water})/\mu_{water}$, where $\mu_{water}=0.202527 cm^{-1}$) \cite{crawford2013research,karimi2015metal}. Thresholded mask results were eroded and then dilated with a disk-shaped structuring element of radius $2$ and $4$ pixels respectively to obtain a final metal mask. Erosion was used to remove very small objects, and a small dilation was used to over-segment the metals. Selected sinograms were input to our \emph{Deep}-MAR framework and final reconstructions were used during testing for qualitative analysis of our proposed method.\par

\subsection{CGAN Training} \label{sec:cgan_tr}
To optimize our CGAN sinogram completion network, the original GAN training strategy \cite{goodfellow2014generative} was followed, where alternations between one gradient descent step on $G$ and one step on $D$ were performed, with the exception that for the first $4$ epochs $k=1,...,4$, $6-k$ iterations on $D$ were done for each gradient descent iteration on $G$. Mini-batch stochastic gradient descent with batch size $6$ was used for both simulated data and real data transfer learning. The standard Adam optimizer \cite{kingma2014adam} was used with learning rate of $0.0002$ and momentum parameters $\beta_1 = 0.9$, $\beta=0.999$. The value of the hyper parameter $\lambda=10$ was decided empirically. For simulated data, the CGAN was trained for $25$ epochs on $50{,}000$ examples using the training scheme just described. Real data was used for transfer learning from the simulated data trained network with just $8{,}530$ pairs of real sinograms. We trained our model on the real dataset for $50$ epochs. In order to show effectiveness of transfer learning, we also trained a model without pre-training on the simulated dataset, i.e., training the model from scratch on the real dataset. At test time, the trained generator network $G$ was used to perform sinogram completion. The network was implemented in Tensorflow \cite{abadi2016tensorflow} heavily borrowing code from the Tensorflow implementation of CGAN\footnote{\texttt{http://github.com/yenchenlin/pix2pix-tensorflow}}.\par

%\begin{figure}[tb]
%	\centering
%	\includegraphics[width=0.4\textwidth]{train_test_mar_cgan_v7_sim.png}
%	\caption{Contribution of both the $\ell_2$ and discriminator loss terms are being presented, overall training loss has very high contribution from $\ell_2$ loss term.}
%	\label{fig:training_loss}
%\end{figure}

\section{Results and Discussion} \label{res}
In this section \emph{Deep}-MAR results are shown and its performance is compared to similar computationally light sinogram completion approaches suitable for use in practical settings. Comparison is made to LI-MAR \cite{kalender1987reduction}, and WNN-MAR \cite{mahnken2003new} on both simulated and real data. For a fair comparison, we set the weight of the original data $\mu_{\text{WNN}}=0$ for WNN-MAR, and use default values for the rest of the parameters suggested in \cite{mahnken2003new}. The performance of our trained CGAN is also analyzed using attention maps and latent space analysis. The performance of our generator network architecture is compared to a popular CNN architecture VDSR \cite{kim2016accurate} in Supplementary Material. This comparison demonstrates the effectiveness of U-net like architectures as generator network. Furthermore, we present additional experimental results in the Supplementary Material. \par

\begin{figure}[tb]
	\begin{center}
	\footnotesize
	\rotatebox{90}{\hspace{2.8em} \emph{Deep}-MAR (Ours)\hspace{6.7em} WNN-MAR \cite{mahnken2003new} \hspace{7.3em} LI-MAR \cite{kalender1987reduction}\hspace{8.3em} Uncorrected}
\includegraphics[width = 0.46\textwidth]{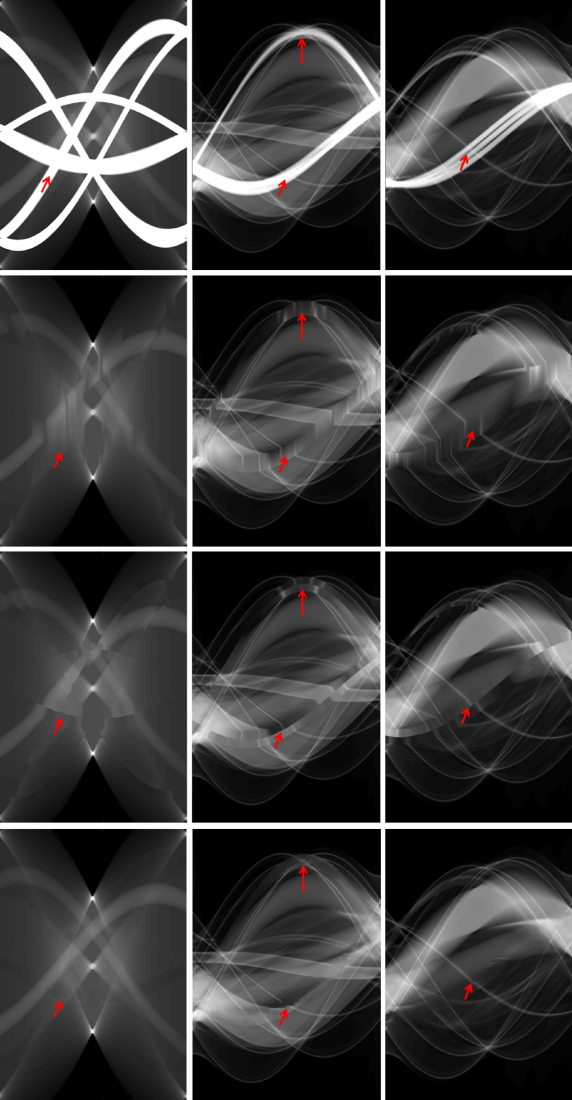}
\end{center}
\footnotesize
\hspace{6.5em}(a) \hspace{8em}(b)  \hspace{8em} (c)
	\caption{Sinogram completion results for a (a) simulated and (b), (c) real data examples using different methods are presented on display window $[0, 5]$ and $[0, 7]$ respectively . First row shows uncorrected sinograms, second row shows the results of LI-MAR, third row shows the results of WNN-MAR, and fourth row shows the results of the proposed \emph{Deep}-MAR approach, showing its ability to accurately complete the sinograms. Red arrows point at some of the regions in sinograms where other methods visibly struggle.  }
	\label{fig:Sino}
\end{figure}

\subsection{Sinogram Completion Experiments}
\begin{figure*}[tb]
	\begin{center}
		\includegraphics[width = 0.99\textwidth]{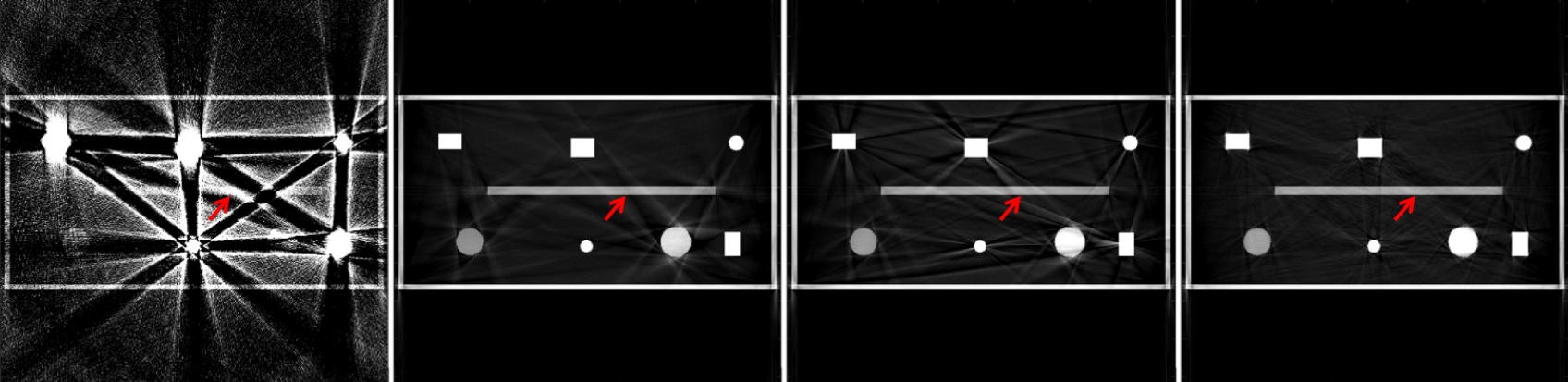}
	\end{center}
	\footnotesize
	\hspace{5em}(a) Uncorrected\hspace{9em}(b) LI-MAR \cite{kalender1987reduction} \hspace{9em}WNN-MAR \cite{mahnken2003new}\hspace{8em}(d) \textbf{\emph{Deep}-MAR (Ours)}
	\caption{MAR reconstruction results for the simulated case of Figure~\ref{fig:Sino} presented on display window $[0, 0.15] \text{ cm}^{-1}$. Column (a) shows results without MAR, column (b) shows the results of LI-MAR, column (c) shows the results of WNN-MAR, and column (d) shows the results of the proposed \emph{Deep}-MAR approach, showing its ability to reduce artifacts.}
	\label{fig:simRec}
\end{figure*}

\begin{figure*}[tb]
\begin{center}
\includegraphics[width = 0.99\textwidth]{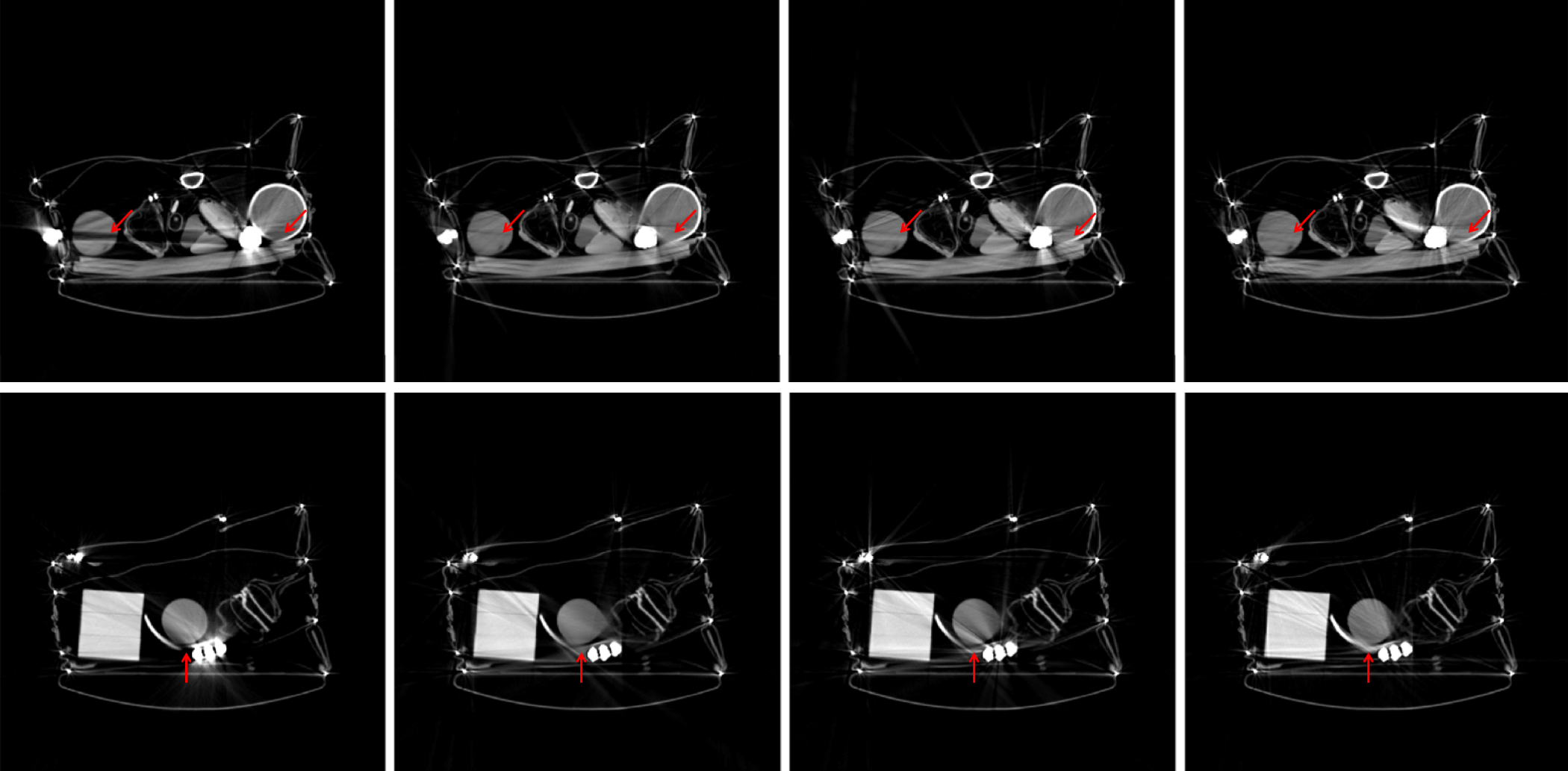}
\end{center}
\footnotesize
\hspace{5em}(a) Uncorrected\hspace{9em}(b) LI-MAR \cite{kalender1987reduction} \hspace{9em}WNN-MAR \cite{mahnken2003new}\hspace{8em}(d) \textbf{\emph{Deep}-MAR (Ours)}
\caption{Reconstruction results for two slices of real data with and without MAR techniques are presented on display window $[0, 0.4] \text{ cm}^{-1}$. Example slices represent a variety in number, and type of materials, and varied level of clutter in the field of view. \emph{Deep}-MAR can reliably recover structure and suppress metal artifacts in these challenging scenarios.}
\label{fig:realRec}
\end{figure*}

Since a key element of our proposed approach is sinogram completion we first focus on the ability of \emph{Deep}-MAR to successfully perform this task. Qualitative sinogram completion results on simulated data are shown in Figure~\ref{fig:Sino}. Each row corresponds to a different sinogram completion case. The first row shows the original uncorrected sinograms, the second row shows the results obtained using LI-MAR \cite{kalender1987reduction}, the third row shows the results obtained using WNN-MAR \cite{mahnken2003new}, and the fourth row shows the results obtained using the proposed learning-based completion approach \emph{Deep}-MAR. Column (a) in Figure~\ref{fig:Sino} corresponds to a simulated example, while column (b) and (c) correspond to two different real data examples obtained from the ALERT TO3 dataset \cite{crawford2013research}. The reference metal-free sinogram for the simulated data example in column (a) is presented in Supplementary Material. The images corresponding to these cases are shown in Figures~\ref{fig:simRec} and \ref{fig:realRec}. LI-MAR and WNN-MAR use limited neighboring data in $1$-D and $2$-D respectively coupled with spatially invariant weights to produce sinogram completion results. While this limited information is adequate in simple cases with few, small metallic objects, these methods struggle in more realistic security scenarios containing larger areas of missing data and complex configurations. As highlighted by the red arrows, the proposed \emph{Deep}-MAR approach produces sinograms with fewer visual artifacts, such as discontinuities at metal trace edges compared to methods widely used in practice. Additional sinogram completion results presented in Supplementary material show that the proposed \emph{Deep}-MAR based completion results are closer to the ground truth sinograms compared to LI-MAR and WNN-MAR completed sinograms.

\begin{table}[tb]
  \centering
  \caption{Sinogram completion performance comparison in terms of average MSE.}
    \begin{tabular}{|l|c|c|c|c|}
    \hline
     & \multicolumn{1}{|p{1cm}}{\centering ~ \\ LI-MAR \\~ }
    & \multicolumn{1}{|p{1.45cm}}{\centering ~ \\ WNN-MAR \\~ }
    & \multicolumn{1}{|p{1.45cm}}{\centering \emph{Deep}-MAR \\ \scriptsize
 (Training from scratch)}
    & \multicolumn{1}{|p{1.5cm}|}{\centering \emph{Deep}-MAR \\ \scriptsize (Transfer Learning)} 
    \\ \hline
	Simulated 	& $0.1029$ 	& $0.0872 $	& - &  $\mathbf{0.0117}$ \\ \hline
	Real 		& $0.0216$ 	& $0.0173$ & $0.0062$ & $\mathbf{0.0043}$\\
\hline
    \end{tabular}%
  \label{tab:mse_sino}%
\end{table}%

To perform a quantitative analysis, we use $510$ test examples from the simulated dataset and $159$ examples from the real dataset prepared for quantitative analysis. We compare sinogram completion results to reference metal-free sinograms using average mean square error (MSE) and present results in Table \ref{tab:mse_sino}. Our sinogram completion method greatly reduces the average MSE as compared to LI-MAR and WNN-MAR: $89\%$ and $87\%$ average relative MSE reduction in simulated data, and $80\%$ and $75\%$ average relative MSE reduction in real data experiments respectively. We also compare performance of our models trained from scratch and using transfer learning on the real dataset using sinogram completion results. The advantage of transfer learning is evident through a significant improvement of $31\%$ in terms of average relative MSE as compared to the model trained from scratch.

\subsection{Simulated Data Reconstruction Experiments}

In this section MAR reconstruction results on simulated data are presented. We use oracle information to segment metals in the simulated data experiments. In Figure~\ref{fig:simRec} a qualitative example is shown corresponding to the sinogram case in Figure~\ref{fig:Sino} (a). This example contains the equivalent of so called bulk and sheet objects which are present in security applications. Column (a) presents results with standard FBP without using any MAR techniques. Other results are produced by applying FBP after performing sinogram completion using the associated sinogram completion method and then inserting the metal objects back into the corresponding reconstructions. Column (b) shows the results of LI-MAR, column (c) shows the results of WNN-MAR, and column (d) shows the results of our new \emph{Deep}-MAR approach. The reference metal-free reconstruction for this example is presented in Supplementary Material. The conventional FBP image naturally exhibits severe streaking artifacts which significantly disrupt both the bulk and sheet objects and which would be undesirable in a security context. While the LI-MAR and WNN-MAR methods have greatly reduced the artifacts, as they are intended to, these methods still exhibit significant distortions which could affect segmentations of the sheet object and density estimation of the bulk objects. The proposed \emph{Deep}-MAR method successfully suppresses nearly all of the streaking artifacts and successfully preserves the sheet object. Additional reconstruction results for simulated data are presented in the Supplementary Material and show the potential of the \emph{Deep}-MAR method

\begin{table}[tb]
  \centering
  \caption{Reconstruction performance comparison of different methods on simulated data in terms of average MSE, SSIM, and PSNR in $ cm^{-1}$ units.}
    \begin{tabular}{|l|c|c|c|c|}
    \hline
  & Uncorrected & LI-MAR & WNN-MAR & \emph{Deep}-MAR\\ \hline
MSE & $3.9 \times 10^{-2}$ & $4.7 \times 10^{-4}$ & $5.5\times 10^{-4}$ & $\mathbf{1.2\times 10^{-4}}$ \\
    \hline
SSIM   & $0.7205$ & $0.9750$ & $0.9713$ & $\mathbf{0.9941}$ \\
    \hline
PSNR   & $33.72$ & $54.67$ & $53.81$ & $\mathbf{60.32}$ \\
    \hline
    \end{tabular}%
  \label{tab:mse_sim}%
\end{table}%

In addition to qualitative comparison, quantitative analysis of the reconstruction results using average MSE, average structural similarity (SSIM), and average peak signal to noise ratio (PSNR) of $510$ simulated test images is presented in Table~\ref{tab:mse_sim}. The proposed \emph{Deep}-MAR method results in significant image quality improvement which is quantified by a $74\%$ and $78\%$ decrease in average MSE as compared to LI-MAR and WNN-MAR methods, respectively. Improvements can be noticed in SSIM and PSNR metrics results as well. Thus the \emph{Deep}-MAR approach not only performs well in qualitatively suppressing metal artifacts, but is also good at correcting the underlying attenuation values, and correct attenuation numbers are important for the successful functioning of security algorithms. \par

\subsection{Real Data Experiments}

In this section end-to-end MAR reconstruction results on real data are presented. First qualitative results for two challenging real data examples are presented in Figure~\ref{fig:realRec}. These reconstructions correspond to the sinograms in Figure~\ref{fig:Sino} (b) and (c). Column (a) presents uncorrected results with standard FBP without using any MAR techniques. Results using LI-MAR, WNN-MAR, and our \emph{Deep}-MAR are presented in Columns (b), (c), and (d) respectively. Red arrows show regions of particular interest. These include large metal objects in proximity to thin sheets or the boundaries of containers (bulk items), as well as interior regions of objects that should be uniform in intensity. In the conventional FBP image it can be seen that the presence of metal causes streaking through the middle of objects which splits them, shading (intensity variation) of what should be homogeneous regions, and the corruption of container boundaries important for segmentation. While conventional MAR methods (LI-MAR and WNN-MAR) help to some extent with streaking, there still exists significant loss of boundary information. Additionally the existing methods produce new streaking artifacts in some cases. In contrast, the \emph{Deep}-MAR method appears to do a remarkable job on this very difficult task of preserving boundary information and object uniformity. The ability to preserve such object information greatly aids subsequent object segmentation and material recognition tasks. Additional reconstruction results for a variety of real data scenes are presented in the Supplementary Material. \par

\begin{table}[tb]
  \centering
  \caption{Reconstruction performance comparison for different methods on real data in terms of average MSE, SSIM, and PSNR in $ cm^{-1}$ units.}
    \begin{tabular}{|l|c|c|c|c|}
    \hline
     & \multicolumn{1}{|p{1cm}}{\centering ~ \\ LI-MAR \\~ }
    & \multicolumn{1}{|p{1.45cm}}{\centering ~ \\ WNN-MAR \\~ }
    & \multicolumn{1}{|p{1.45cm}}{\centering \emph{Deep}-MAR \\ \scriptsize
 (Training from scratch)}
    & \multicolumn{1}{|p{1.5cm}|}{\centering \emph{Deep}-MAR \\ \scriptsize (Transfer Learning)}
    \\ \hline
	MSE 	&  $1.4\times 10^{-4}$ 	& $1.8\times 10^{-4}$ 	& $7.5\times 10^{-5}$ & $\mathbf{5.8\times 10^{-5}}$ \\
    \hline
	SSIM   	&  $0.8158$				& $0.8029$ 				& $0.8605$ & $\mathbf{0.8840}$ \\
    \hline
	PSNR   	&  $30.49$ 				& $29.61$ 				& $32.68$ &  $\mathbf{33.91}$ \\
    \hline
    \end{tabular}%
  \label{tab:mse_real}%
\end{table}%

In addition to qualitative results, quantitative analysis of real data cases was performed on a set of $159$ real data examples (as described in Section \ref{real_testing}). These results are presented in the Table~\ref{tab:mse_real}. This quantitative analysis reveals that our proposed \emph{Deep}-MAR method results in $59\%$ and $68\%$ reduction of average MSE as compared to LI-MAR and WNN-MAR methods, respectively. Improvements in average SSIM and PSNR metrics are also significant. We also compare performance of our models trained from scratch and using transfer learning on the real dataset using reconstructed images. The advantage of transfer learning is evident through a significant improvement of $23\%$ in terms of average relative MSE as compared to the model trained from scratch. \par

Artificial generation of objects can be a concern when using machine learning algorithms in general imaging applications. We have not observed any such artificial objects generated by our CGAN approach, though imperfections at different stages of our method can and do lead to artifacts such as streaking, boundary drop out, and intensity shading. These artifacts are similar to what is seen with other completion-based MAR methods, though we believe the enhanced power of the CGAN approach reduces them.

\begin{figure*}[tb]
\centering
\includegraphics[width = 0.99\textwidth]{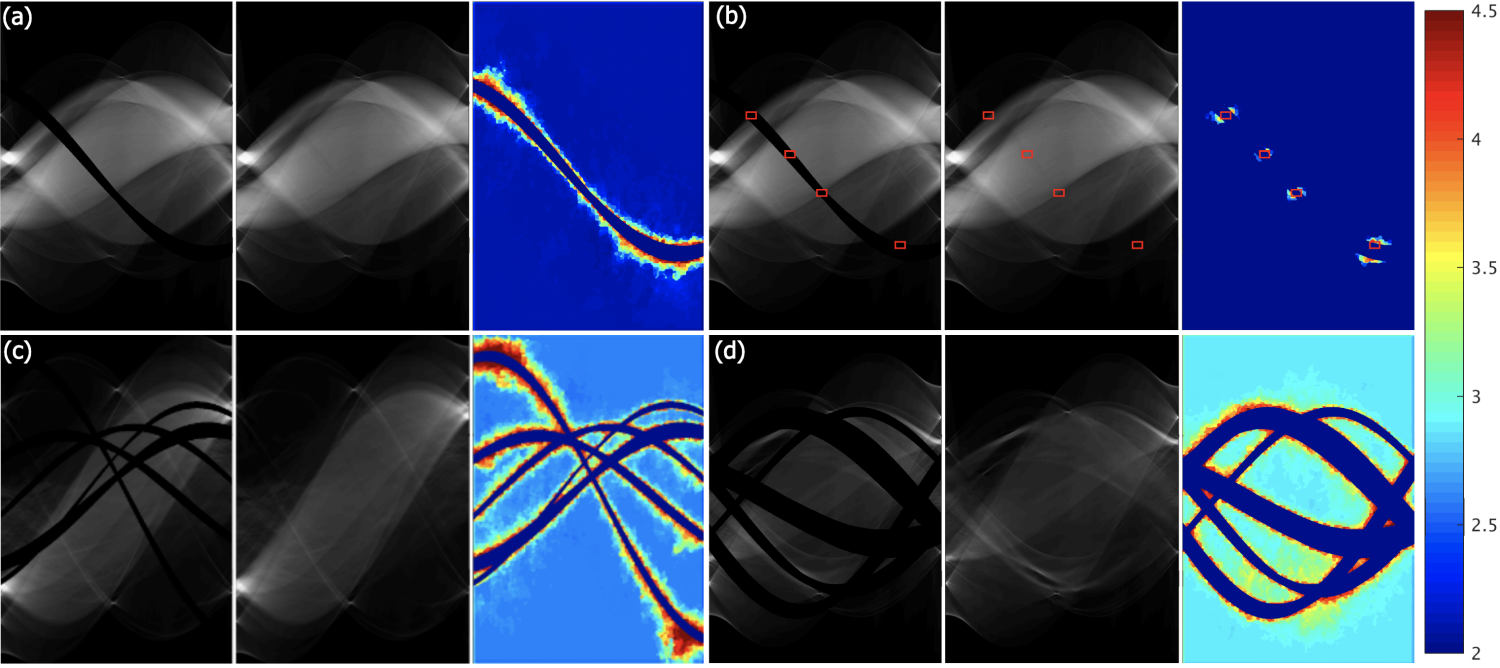}
	\caption{Attention maps for an easy case, two challenging cases, and four ROIs in the easy case are presented. (a) Attention map of an example with one metal object, (b) attention map focused on four ROIs for the same example shown in red, (c), (d) attention maps for two examples with $5$ metal objects. In all cases the missing data sinogram is shown in the left panel, the completed sinogram is show in the middle panel, the attention map is shown in the right panel. It can be seen from the maps that small sinogram gaps seem connected to small decision support regions while challenging scenarios as presented in (d) needs decision support from a larger portion of the sinogram.}
\label{fig:attnMaps}
\end{figure*}

\subsection{Attention Maps}
To gain an enhanced understanding of the behavior of our deep network approach we extend the occlusion sensitivity analysis work of Zeiler and Fergus \cite{zeiler2014visualizing} to \emph{image-to-image} translation tasks. The work in \cite{zeiler2014visualizing} focused on image classification tasks and aimed to compute attention maps by occluding different regions of the input image and examining the resulting impact on performance of the deep network on the corresponding image classification task. The idea is that occlusion of more important image areas to the network will correspondingly have greater resulting performance impact. In this work, we apply this idea by repeatedly replacing an $11\times11$ patch of sinogram values with random data sampled from a uniform distribution \emph{U}$(0,m)$ (which is thus uninformative) and then examining the resulting sinogram completion performance. Here, $m$ is the maximum value in the sinogram under study. We use overlapping patches with a sliding distance of $6$ pixels. We then compute the corresponding MSE for the sinogram regions completed by learned network and generate a $2$-D attention map by associating this MSE value with the center of the corresponding noise patch. For better scaling, we display $\log_{10}(\cdot)$ of the MSE. Larger values in this resulting map should correspond to regions of the sinogram with more important influence on the solution generated by the network.\par 

Attention maps generated using this strategy are presented in Figure~\ref{fig:attnMaps}. These attention maps provide us a way to understand which areas of the sinogram the generator network considers most important to the result. Figure~\ref{fig:attnMaps} (a) presents the attention map for an example with one metal object. Figures~\ref{fig:attnMaps} (c) and (d) present attention maps for two challenging examples with $5$ metal objects. Figure~\ref{fig:attnMaps} (b) presents attention maps for selected regions of interest (ROIs) shown in red. Figures~\ref{fig:attnMaps} (a) and (c) show that the pixels closest to the missing data appear to be the most important in generating an estimate, which seems logical. They also show that the size of the region of major influence also appears to grow with the width of the region of missing data. Figure~\ref{fig:attnMaps} (d) shows that for challenging problems with significant missing data, more and more of the sinogram is used to generate an estimate. This ``adaptivity" is in contrast to existing LI-MAR, and WNN-MAR methods where the region of influence is fixed and unchanging. 

\begin{figure*}[tb]
\centering
\subfigure[Input projection data space]{\includegraphics[height=0.19\textheight, cfbox=blue 0.5pt 0.5pt]{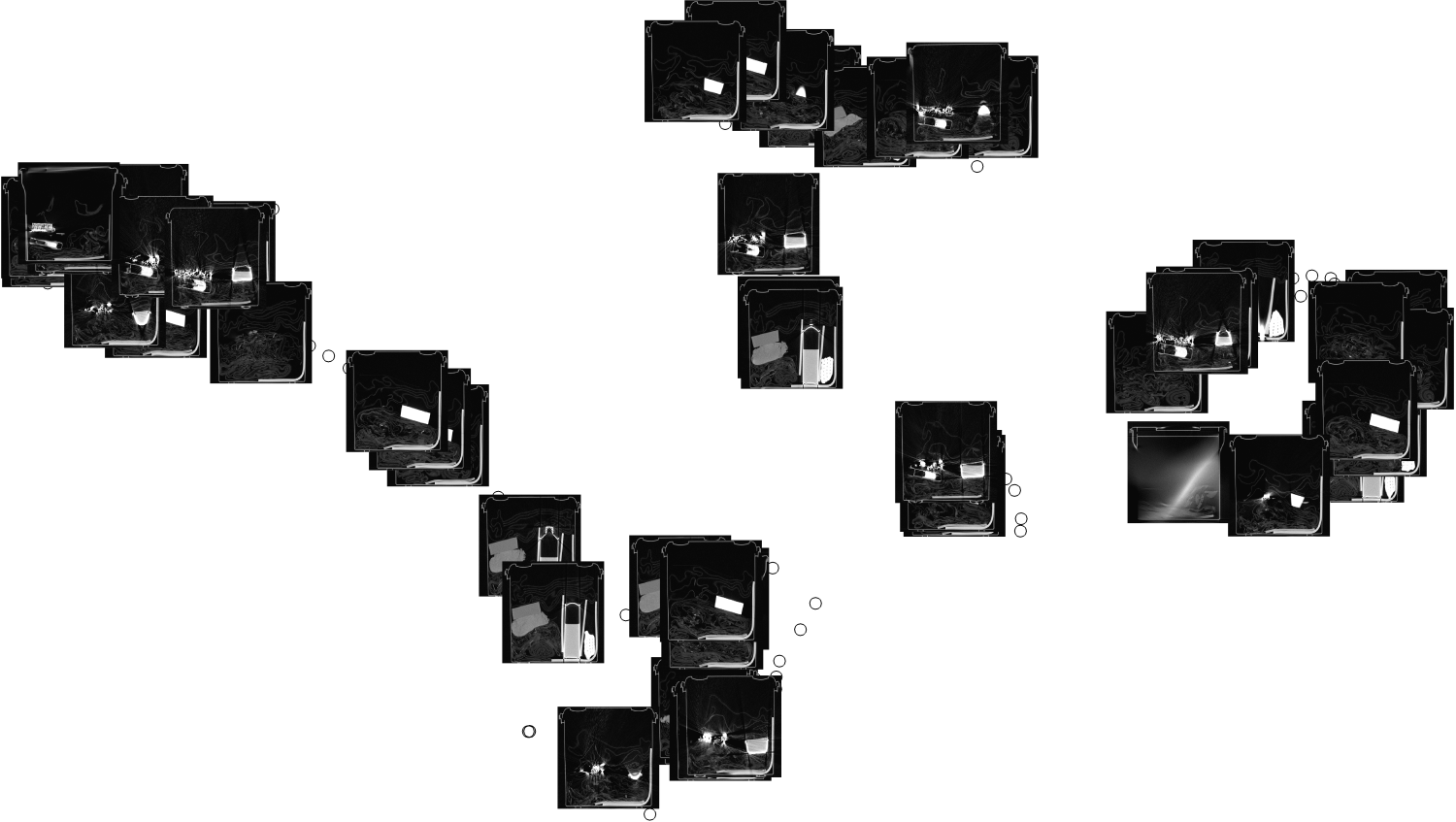}} \hspace{0.1em}
\subfigure[Latent representation space]{\includegraphics[height=0.19\textheight, cfbox=blue 0.5pt 0.5pt]{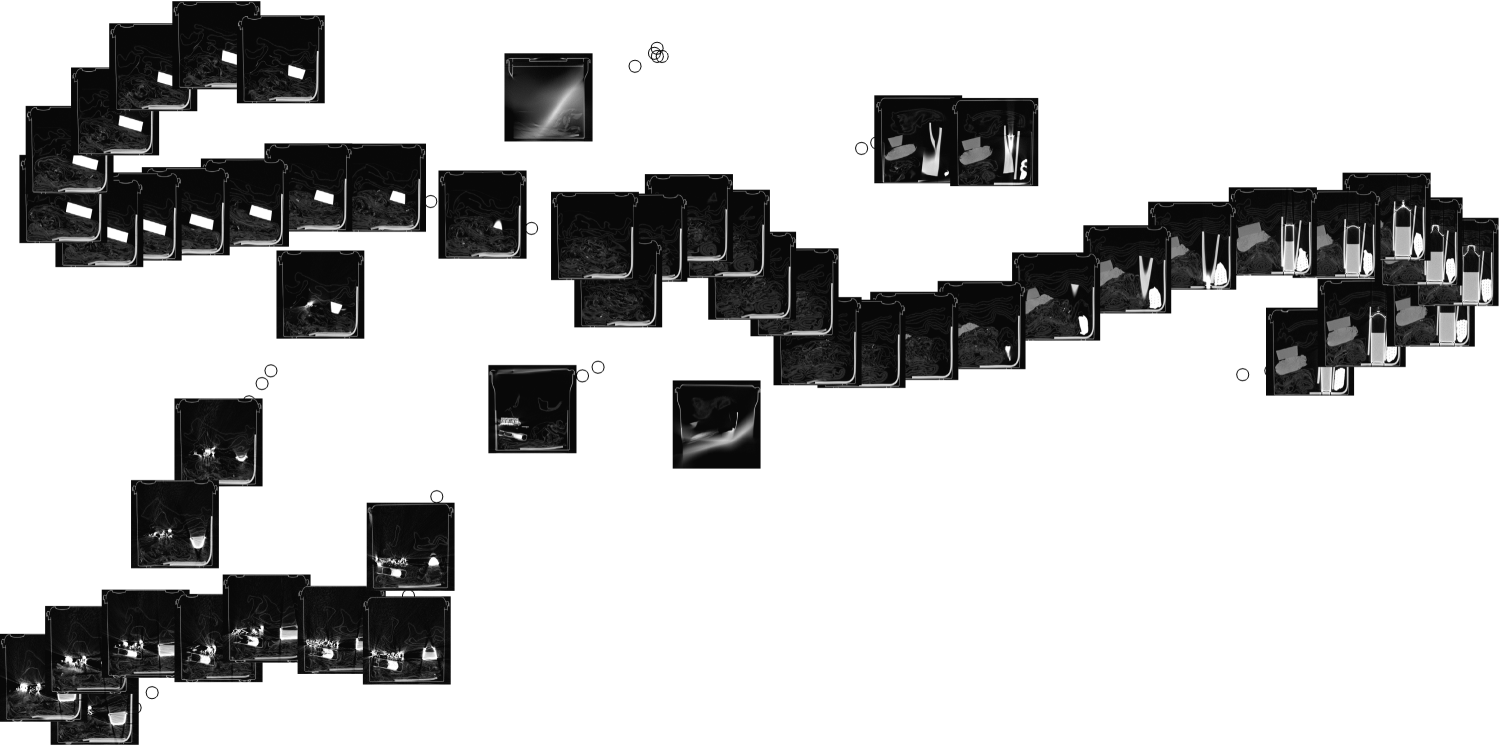}}
	\caption{A $2$-D t-SNE visualization of (a) input sinogram data and (b) corresponding latent representation learned by CGAN encoder are presented alongwith the corresponding reconstruction images. The t-SNE visualization of the input sinogram does not exhibit any structure, i.e., similar image slices are not assigned similar t-SNE representations. However, the t-SNE visualization of our CGAN encoder's learned latent representation shows that similar image slices are assigned similar representations. This exhibited structure suggests that our CGAN encoder has successfully learned the semantic information in the sinograms.}
\label{fig:tsne}
\end{figure*}

\subsection{Latent space analysis}

The ability to learn latent structure could be argued to be one of the reasons behind the success of deep networks. The generator network that performs sinogram completion in our CGAN framework follows an encoder-decoder architecture, where the encoder maps high-dimensional input data to a low-dimensional latent representation, which is then used by the decoder to create the final image. One question is whether the resulting learned latent space representation is semantically meaningful -- that is, whether it is actually capturing underlying relationships between different scene configurations. To investigate this question we applied the popular t-SNE \cite{maaten2008visualizing} dimensionality reduction technique to visualize the high-dimensional input and corresponding learned latent space representations of various examples. We used sinogram data from all the 2-D slices of a suitcase and applied the t-SNE approach to the input sinogram data as well as the encoder produced latent sinogram representation to compare the representational information in these two domains. A $2$-D presentation was created wherein each point in the input and latent space was tagged with its corresponding reconstructed image. Of course, similar images will have similar corresponding sinograms. The results are presented in Figure ~\ref{fig:tsne}. The images associated with the input sinogram data do not appear to exhibit significant clustered structure. It does not appear that similar images/slices are grouped together in the original space. In contrast, similar image slices \emph{do} appear to group naturally together in the latent space representation case. This result suggests that our CGAN encoder is able to learn a semantically meaningful latent representation from the sinogram projection data. \par

\section{Conclusion \label{conc}}

Metal artifacts pose serious challenges to interpretability of CT reconstructions. While many sinogram completion based methods and image post-processing methods have been proposed, the problem is still very challenging and the subject of active research. In this paper, we present \emph{Deep}-MAR, a new DL-based framework for data domain MAR, that uses a CGAN for data domain sinogram completion. Additionally, we demonstrate the successful use of transfer learning from simulated training data to a small set of real data. Excellent performance is achieved on challenging real security examples and the method is computationally light enough to be used in practical situations. We evaluate the performance of our method on simulated and real datasets and compare the results to two popular MAR methods: LI-MAR, and WNN-MAR. Qualitative and quantitative analysis of reconstruction results show the potential of our approach over conventional methods. In particular, the proposed method robustly suppresses metal artifacts and retains significant structure and reconstructed density numbers. Although, we consider a security scenario in this paper, our technique is generic in that it can be trained and applied to reduce metal artifacts in other applications as well, such as medical imaging. 
 
\ifCLASSOPTIONcaptionsoff
  \newpage
\fi

\small
\bibliographystyle{IEEEbib}
\bibliography{literature.bib}

 \begin{IEEEbiography}{Muhammad Usman Ghani}
 received the B.S. degree in Electrical Engineering from COMSATS Institute of Information Technology, Lahore, Pakistan, in $2013$, and the M.S. degree in Computer Science from Sabanci University, Istanbul, Turkey, in $2016$. He is currently a Ph.D. candidate at the Department of Electrical and Computer Engineering, Boston University, Boston, MA, USA. His research interests include machine learning, computational imaging, computer vision, and biomedical image processing. 
 \end{IEEEbiography}

\begin{IEEEbiography}{W. Clem Karl}
 (M'91–SM'00–F'14) received the
 Ph.D. degree in electrical engineering and computer
 science from the Massachusetts Institute of Technology.
 He is currently the Chair of the Electrical
 and Computer Engineering Department and a
 member of the Biomedical Engineering Department
 with Boston University. His research interests are in
 the areas of statistical signal and image processing,
 estimation, detection, and medical signal and image
 processing. His IEEE activities include the inaugural
 Editor-in-Chief of the IEEE TRANSACTIONS ON
 COMPUTATIONAL IMAGING (2014-2017), the IEEE Publication Services
 and Products Board Strategic Planning Committee (2015-Present), the Editor-in-
 Chief of the IEEE TRANSACTIONS ON IMAGE PROCESSING (2013-2014),
 the SPS Nominations and Appointments Committee (2014), a member of
 the SPS Publications Board (2012-Present), a member of the IEEE SPS
 Conference Board (2012-Present), a Member-at-Large of the IEEE SPS Board
 of Governors (2011-2013), a member of the IEEE TRANSACTIONS ON
 MEDICAL IMAGING Steering Committee (2011-2013), the Vice Chair of the
 SPS Bio Imaging and Signal Processing Technical Committee (2009-2010),
 a member of the SPS Bio Imaging and Signal Processing Technical Committee
 (2007-2013), a member of the IEEE International Symposium on Biomedical
 Imaging Steering Committee (2009-2010), the General Chair of the 2009
 IEEE International Symposium on Biomedical Imaging, and a member of
 the SPS Image, Video, and Multidimensional Signal Processing Technical
 Committee (2003-09). He was a Co-Organizer of the Special Session of the
 2012 IEEE Statistical Signal Processing Workshop on Challenges in
 High-Dimensional Learning (2011-2012), a Co-Organizer of the Special
 Session of the 2012 IEEE Statistical Signal Processing Workshop on Statistical
 Signal Processing and the Engineering of Materials (2011-2012), a Technical
 Program Committee of the 2012 IEEE Statistical Signal Processing
 Workshop (2012), an Americas Liaison of the 2012 IEEE International Symposium
 on Biomedical Imaging, an Organizer of the Workshop on Large Data
 Sets in Medical Informatics Part of the Institute for Mathematics and its Applications
 Thematic Year on the Mathematics of Information (2010-2011), and
 a Program Committee of the IS\&T/SPIE Computational Imaging Conference
 (2012-present).
 \end{IEEEbiography}

\vfill
% that's all folks
\end{document}

% \linenumbers

\title{Fast Enhanced CT Metal Artifact Reduction using Data Domain Deep Learning -- Supplementary Material}

\author{Muhammad Usman Ghani, %~\IEEEmembership{Member,~IEEE,}
        W. Clem Karl,~\IEEEmembership{Fellow,~IEEE} 
        \thanks{This material is based upon work supported by the U.S. Department of Homeland Security, Science and Technology Directorate, Office of University Programs, under Grant Award 2013-ST-061-ED0001. The views and conclusions contained in this document are those of the authors and should not be interpreted as necessarily representing the official policies, either expressed or implied, of the U.S. Department of Homeland Security.}
\thanks{M. U. Ghani and W. C. Karl are with the Department
of Electrical and Computer Engineering, Boston University, Boston,
MA, 02215 USA (e-mail: \texttt{\{mughani, wckarl\}@bu.edu)}.}}

\markboth{IEEE Transactions on Computational Imaging}%
{Ghani \MakeLowercase{\etal}: Fast Enhanced CT Metal Artifact Reduction using Data Domain Deep Learning -- Supplementary Material}

\maketitle
This supplement provides additional results and information in support of our paper \cite{ghani2019deepmar}. The rest of this supplement is structured as follows. Section~\ref{supp:refimage} provides the reference results for the simulated example considered in the paper. Section~\ref{supp:loss} contains information about the value in using an adversarial discriminator loss. We compare performance of our modified U-Net architecture to a popular network, VDSR \cite{kim2016accurate}, for sinogram completion task in Section \ref{supp:network_arch_comp}. Sinogram completion and reconstruction results for additional simulated and real data examples are presented and discussed in Section~\ref{supp:results}. The metallic and non-metallic materials used in the simulated data experiments are listed in Section~\ref{supp:materials}.

 %%[[WCK: The supplementary material needs a better "introduction" that gives a quick outline of everything in there. 
 %% Right now you start with figure 1, then provide an overview of section S3 and S4, 
 %% but there is no mention of S1 or S2!]]
 
 %%[[MUG: Thanks, it looks better now! ]

\section{Reference Images for Simulated Example}
\label{supp:refimage}

\begin{figure}[tb]
	\centering
	\includegraphics[height=0.15\textheight]{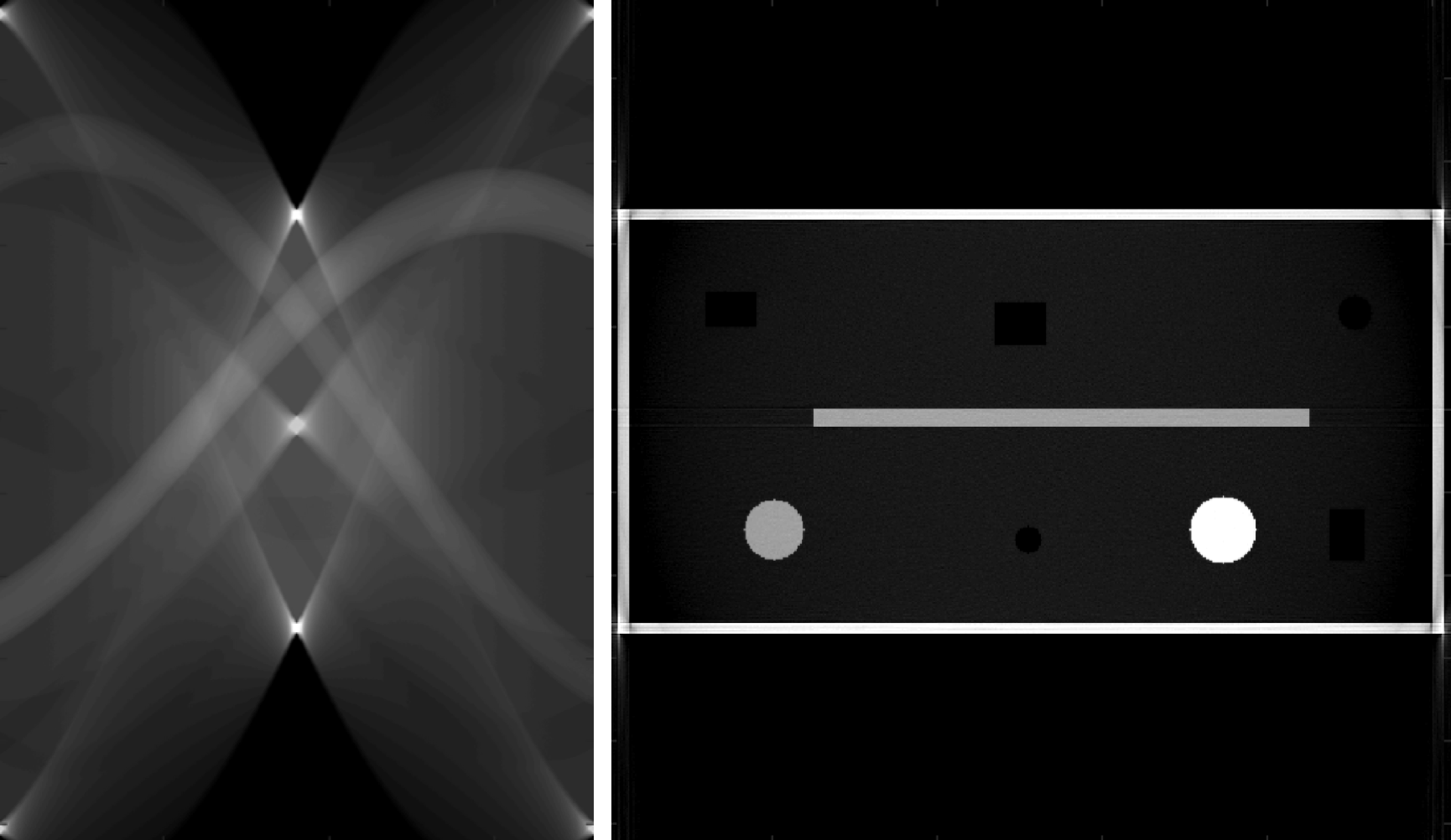}\\
	\footnotesize
	\hspace{1em} (a) Sinogram \hspace{4em} (b) Reconstruction \hspace{2em}
	\caption{Reference sinogram and reconstructed image without any metal for simulated example considered in our manuscript \cite{ghani2019deepmar} are presented.}
	\label{fig:ref_sim_supp}
\end{figure}

The reference metal-free sinogram and reconstructed image for the simulated example considered in Figure $5$ and Figure $6$ of our manuscript \cite{ghani2019deepmar} are presented in Figure \ref{fig:ref_sim_supp}.

\section{Contribution of the $\ell_2$ and discriminator loss terms} 
\label{supp:loss}

\begin{figure}[tb]
	\centering
	\includegraphics[width=0.4\textwidth]{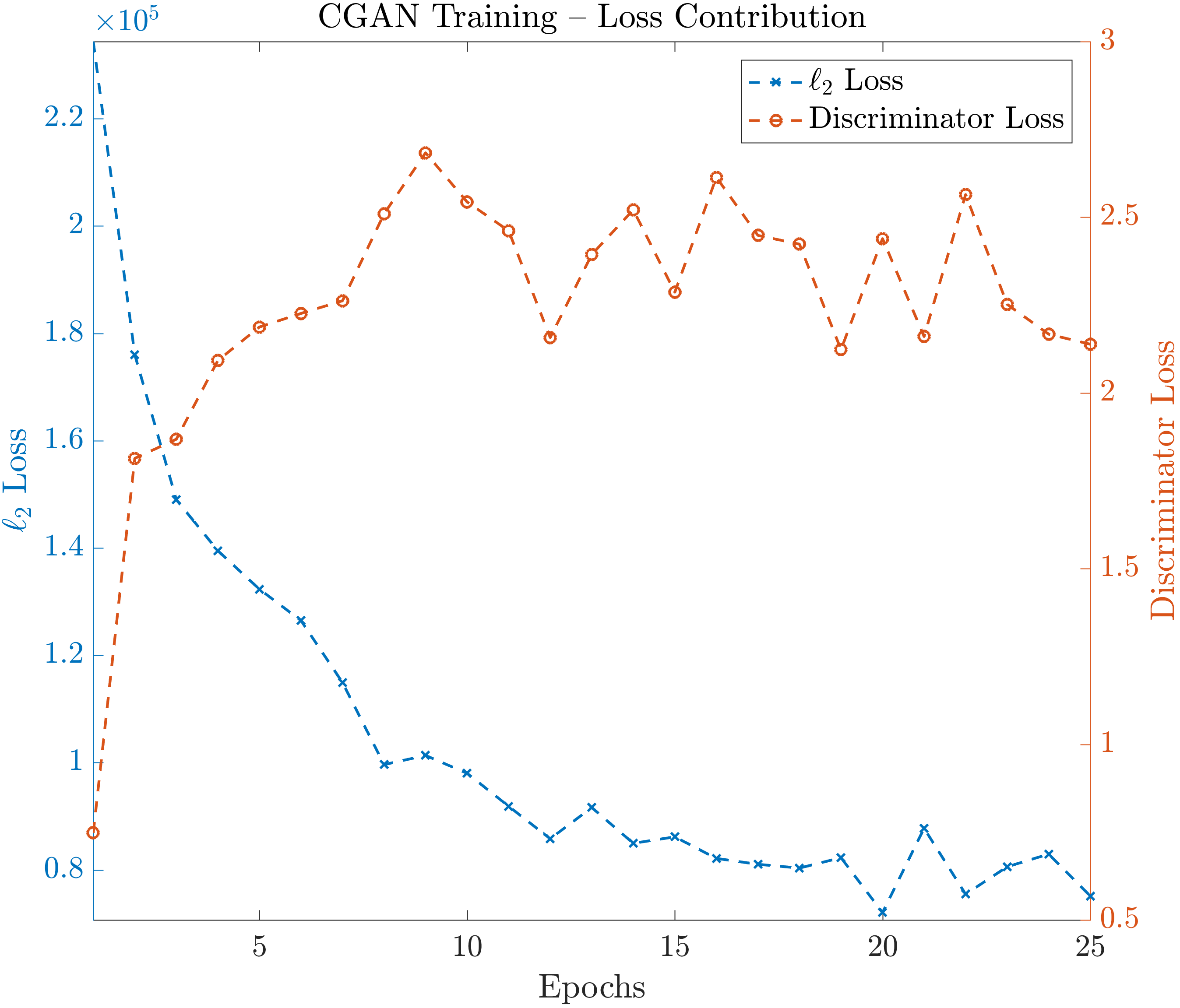}
	\caption{Contribution of both the $\ell_2$ and discriminator loss terms are being presented, overall training loss has very high contribution from $\ell_2$ loss term.}
	\label{fig:training_loss}
\end{figure}

\begin{figure*}[tb]
\centering
\includegraphics[width = 0.99\textwidth]{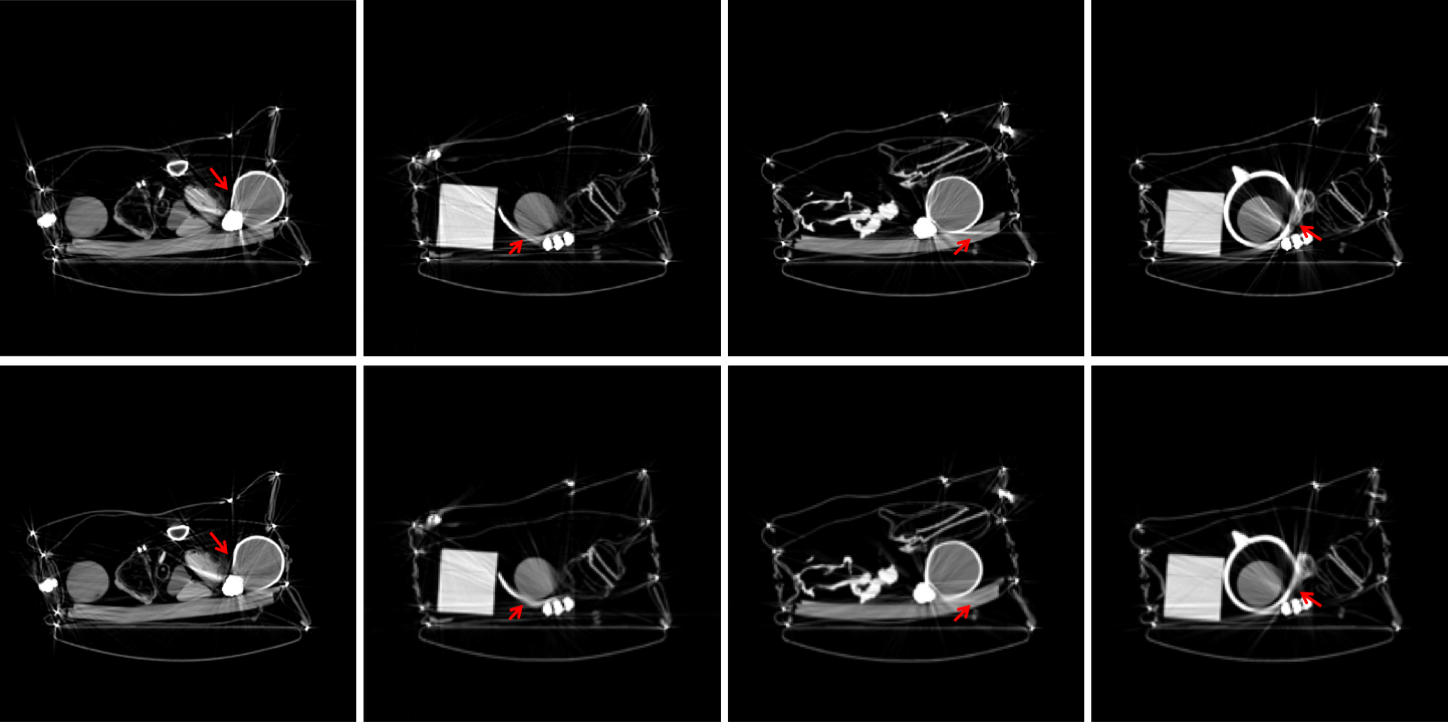}
\footnotesize
\hspace{9em}(a) \hspace{14em} (b) \hspace{14em} (c) \hspace{14em} (d) \hspace{0em}
\caption{Sample results using the final transfer learned CNN and CGAN models are presented. Top row presents results using CNN trained only with the $\ell_2$ loss and bottom row presents results using CGAN trained on a combination of the $\ell_2$ and discriminator loss terms. Results in (a) and (b) correspond to examples considered in Figure $7$ of our paper \cite{ghani2019deepmar}. Results in (c) and (d) correspond to examples considered in Figure \ref{fig:s_realRes1}.} 
\label{fig:cnn_TO3_results}
\end{figure*}

One question is the value of using an adversarial loss over a simpler, conventional $\ell_2$ based CNN for sinogram completion. To understand the relative impact of the two terms the CGAN model was trained on the simulated dataset using the setup described in Section V-B of \cite{ghani2019deepmar} for $25$ epochs. The individual contribution of both the $\ell_2$ and the discriminator loss terms are presented in Figure \ref{fig:training_loss}. The training loss curves suggest that the $\ell_2$ contribution dominates the overall loss. In order to see if including the discriminator loss term offers any performance gains, the sinogram completion network was trained with the same modified U-Net architecture with only the $\ell_2$ loss (the conventional CNN training loss) and the resulting outcomes compared. The sample results obtained from the final transfer learned CNN and CGAN models are presented in Figure~\ref{fig:cnn_TO3_results}. CNN results are presented in the top row and CGAN results are presented in the bottom row. Each column presents results for a different example. It is clear from the presented results that CGAN performs a better job at artifact reduction as compared to CNN. This demonstrate the advantages of using the full CGAN loss (described in Equation $1$ of our paper \cite{ghani2019deepmar}) over the simpler $\ell_2$-based CNN in terms of object boundary preservation and intensity homogeneity, as illustrated by the red arrows in the results. A quantitative comparison of CNN and CGAN results on a real test dataset was also performed and is given in Table~\ref{tab:cnn_cgan_comp}. Significantly better performance using a CGAN at both sinogram and image reconstruction stages can be seen in terms of all considered metrics. 

%%[[WCK: OK since there is no page limit I think you really need to produce the *comparison* in this figure! Right now you make the reader figure out where the CGAN results are (which are spread out) and then compare them. Just put them all in one place (e.g. two rows) to make the comparison easy. Then you can talk about the comparison here. ]]

%%[[MUG: I have added CGAN results in bottom row and updated figure caption and appropriate text in the previous paragraph discussing these results.]

\begin{table}[tb]
  \centering
  \caption{Sinogram completion and image reconstruction performance comparison on real dataset.}
    \begin{tabular}{|l|c|c|}
    \hline
    & \multicolumn{1}{p{1.7cm}|}{\centering Trained with $\ell_2$ loss (CNN)}
    & \multicolumn{1}{p{3.2cm}|}{\centering Trained with $\ell_2$ + discriminator loss (CGAN)} 
    \\ \hline
	MSE (sinogram) & 0.0048 &  $\mathbf{0.0043}$ \\ \hline
	MSE (reconstruction)  	& $6.1 \times 10^{-5}$ & $\mathbf{5.8 \times 10^{-5}}$\\\hline
	SSIM (reconstruction)  	& $0.8776$ & $\mathbf{0.8840}$\\\hline
	PSNR (reconstruction)  	& $33.59$ & $\mathbf{33.91}$\\ \hline
    \end{tabular}%
  \label{tab:cnn_cgan_comp}%
\end{table}%

\section{Network Architecture Comparison} \label{supp:network_arch_comp}

\begin{table}[tb]
  \centering
  \caption{Comparison of VDSR \cite{kim2016accurate} and modified U-Net architectures on simulated data in terms of average Mean Square Error (MSE).}
    \begin{tabular}{|l|l|l|l|}
    \hline
  & Uncorrected & VDSR & Modified U-Net  \\ \hline
Average MSE & $1.3\times 10^{-5}$ & $1.9\times 10^{-6}$ & $5.3\times 10^{-7}$\\
    \hline
    \end{tabular}%
  \label{tab:mse_nets}%
\end{table}%

\begin{figure}[tb]
\centering
\subfigure[VDSR \cite{kim2016accurate}]{\includegraphics[height=0.15\textheight]{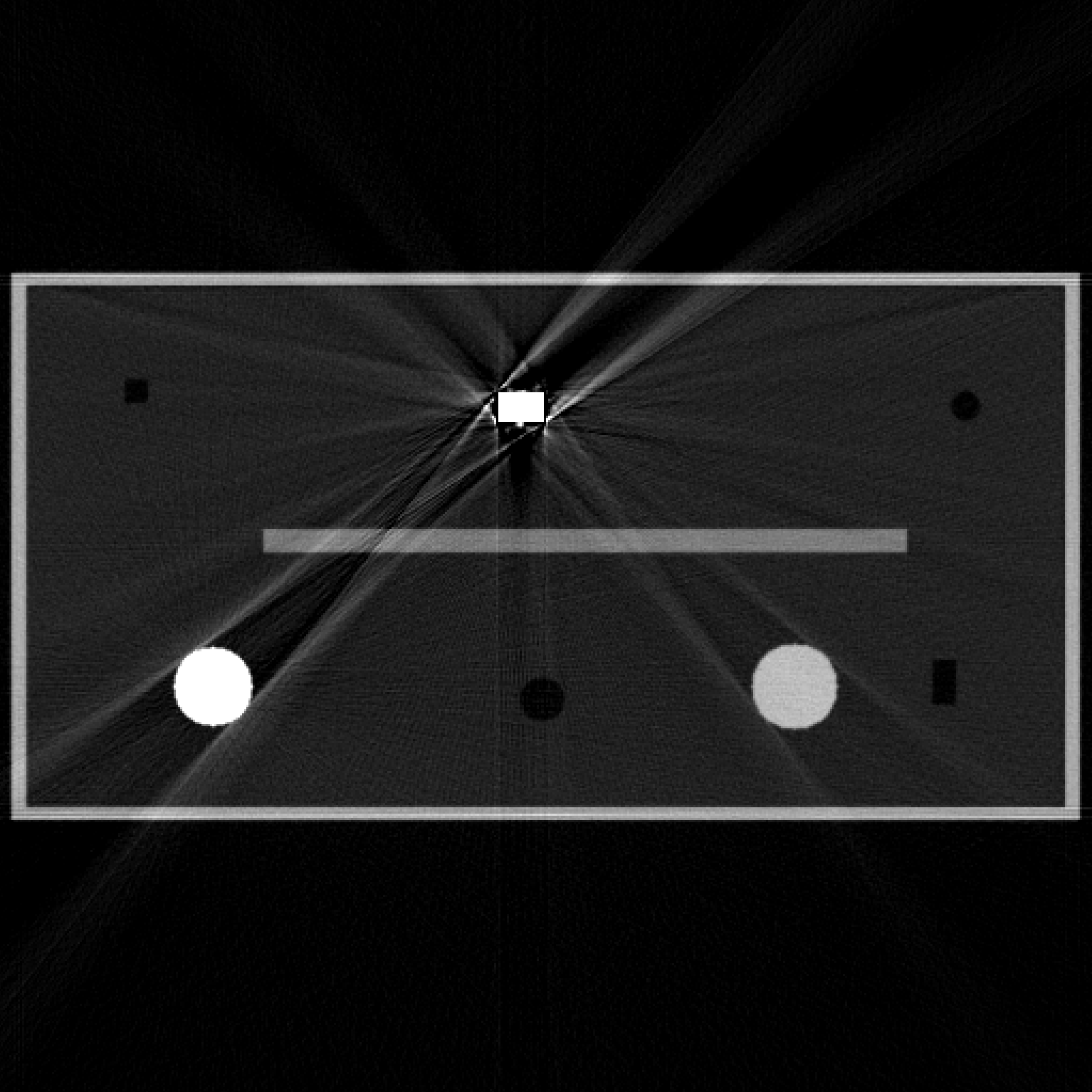}}
\hspace{-5pt} 
\subfigure[Modified U-Net]{\includegraphics[height=0.15\textheight]{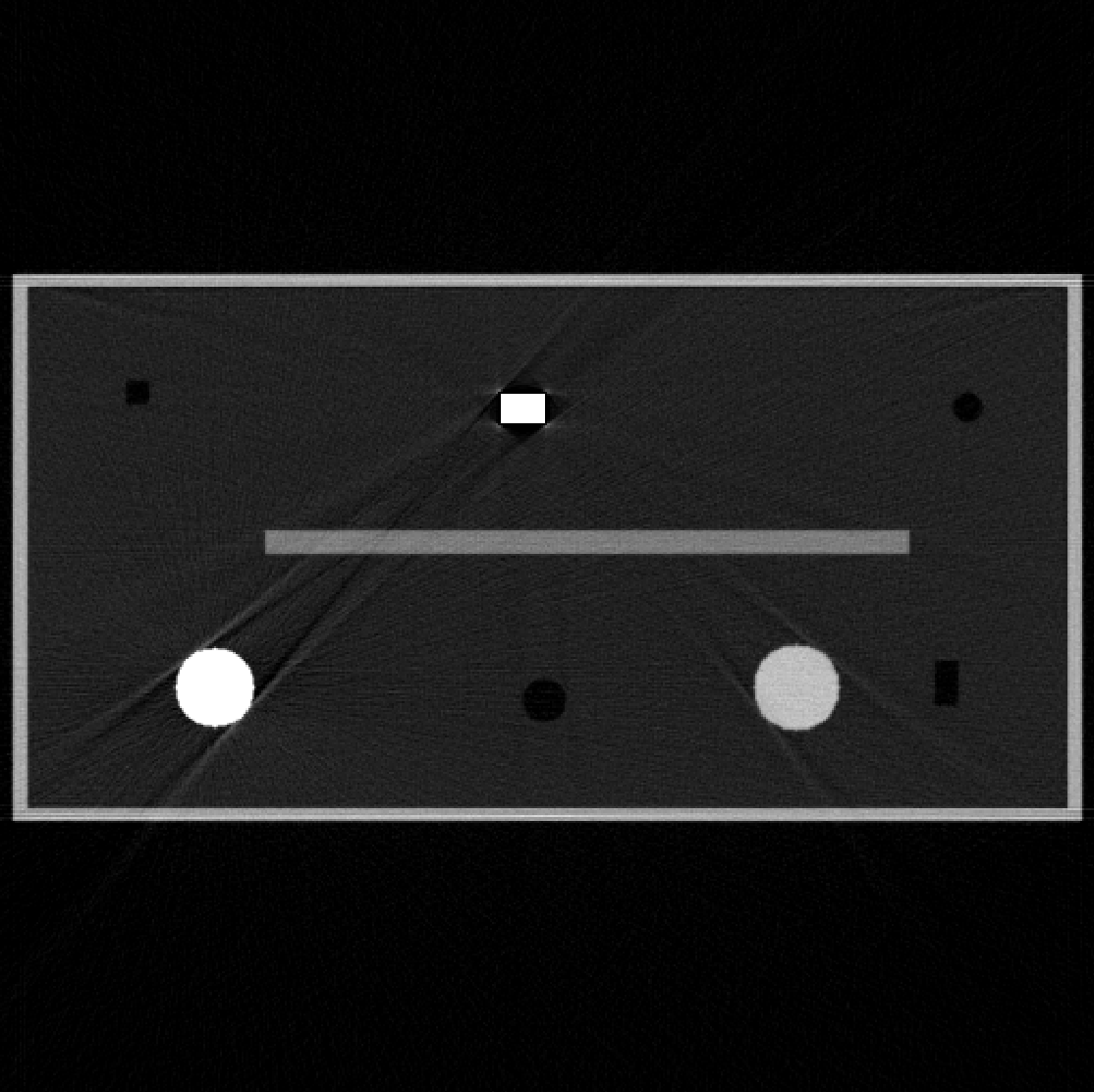}}
\caption{Reconstruction results for an example containing one metallic object using VDSR \cite{kim2016accurate} and Modified U-Net architectures for sinogram completion.}
\label{fig:comp_network_archs}
\end{figure}

In order to show the effectiveness of the U-Net-derived network architecture used in our \emph{Deep}-MAR sinogram completion approach, we compare its performance to a different fully convolutional network architecture, the VDSR \cite{kim2016accurate}, which was proposed for image super-resolution. We only consider the VDSR network here since the VDSR and the U-Net are two important classes of network architectures used in computational imaging research \cite{jin2017deep,lee2017view,zhang2017beyond,meinhardt2017learning,antholzer2018deep,han2018framing,ye2018deep}. These networks reflect different philosophies. The U-Net promotes an idea of multi-scale feature learning while the VDSR reflect same-scale processing resulting in single-scale learned features. VDSR is a $20$ layer fully convolutional network, where each layer contains $64$ kernels of size $3\times3$ with a stride of $1$. We trained VDSR and the modified U-Net architecture (used as the generator network in our paper) on a subset of the simulated dataset containing one metal object. We trained VDSR using the Adam optimizer \cite{kingma2014adam} with $0.1$ learning rate, and the modified U-Net using the same settings as described in \cite{ghani2019deepmar}. We used $\ell_2$-loss to train both networks for $10$ epochs. Training data included $10,000$ examples and test data included $100$ examples. Reconstruction results for one of the test examples using both trained networks are presented in the Figure~\ref{fig:comp_network_archs}. It shows that VDSR achieves only partial artifact suppression, while the modified U-Net architecture-based MAR successfully suppresses most of the streaks and maximally covers the structure lost in streaking artifacts. Further, quantitative results presented in Table~\ref{tab:mse_nets} shows a similar trend in that the modified U-Net reduces the average MSE by $72\%$ compared to VDSR. We believe that the reason behind the excellent performance of the U-Net architecture in this application is that it captures multi-scale information, and the skip connections aid in the transfer of low-level information across the layers \cite{pix2pix17}. In contrast, VDSR captures information at the same scale in all of its layers, which appears to limit its performance for the sinogram completion task. \par

%%[[WCK: Apparently I don't have these figures because latex will not compile for this for me. In the PDF you gave me the figure labels were wrong, so please check and update them! ]]

%%[[MUG: Thanks, apparently latex messes up with figure number if you put the label before caption. ]

\section{Sinogram Completion and Reconstruction Results} \label{supp:results}
\begin{figure*}[tb]
\begin{center}
\includegraphics[width = 0.99\textwidth]{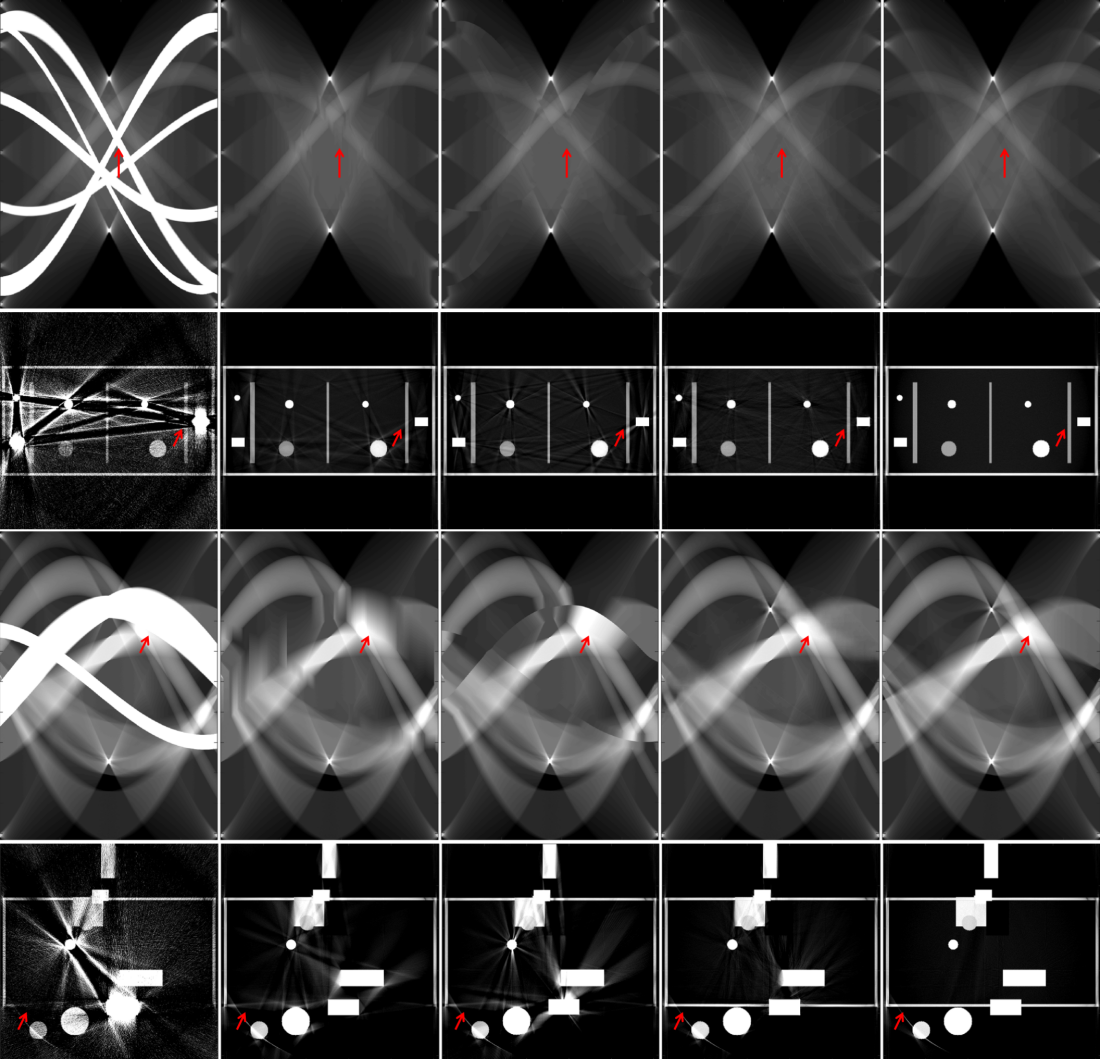}
\end{center}
\footnotesize
\hspace{3em}(a) Without MAR\hspace{5.5em}(b) LI-MAR \cite{kalender1987reduction} \hspace{6em}WNN-MAR \cite{mahnken2003new}\hspace{5.25em}(d) \textbf{\emph{Deep}-MAR (Ours)} \hspace{2em} (e) Reference (metal-free)
\caption{Sinogram completion and reconstruction results for two simulated examples with and without MAR techniques are presented. Sinogram results are presented on display window $[0, 5]$ in first and third rows, and corresponding reconstructions are presented on display window $[0, 0.15] cm^{-1}$ in second and fourth rows. Sinograms and corresponding reconstructions without any MAR processing are presented in column (a), results for LI-MAR, WNN-MAR, and our \emph{Deep}-MAR are presented in column(b), (c), and (d) respectively. Reference metal-free sinograms and corresponding FBP reconstructions without any metals are presented in column (e). Example slices represent a variety in number, and type of materials. \emph{Deep}-MAR can reliably complete the holes in sinogram and therefore corresponding reconstructions recover structure and suppress metal artifacts both scenarios.}
\label{fig:s_sinoRes}
\end{figure*}

Additional sinogram completion and reconstruction results for two simulated examples are presented in Figure~\ref{fig:s_sinoRes}. The sinogram completion results are presented in first and third rows and show that our \emph{Deep}-MAR method produces completed sinograms which are very close to reference metal-free sinograms. Further, reconstruction results presented in second and fourth rows demonstrate great artifact suppression in the \emph{Deep}-MAR results. \par

\begin{figure*}[tb]
\begin{center}
\includegraphics[width = 0.99\textwidth]{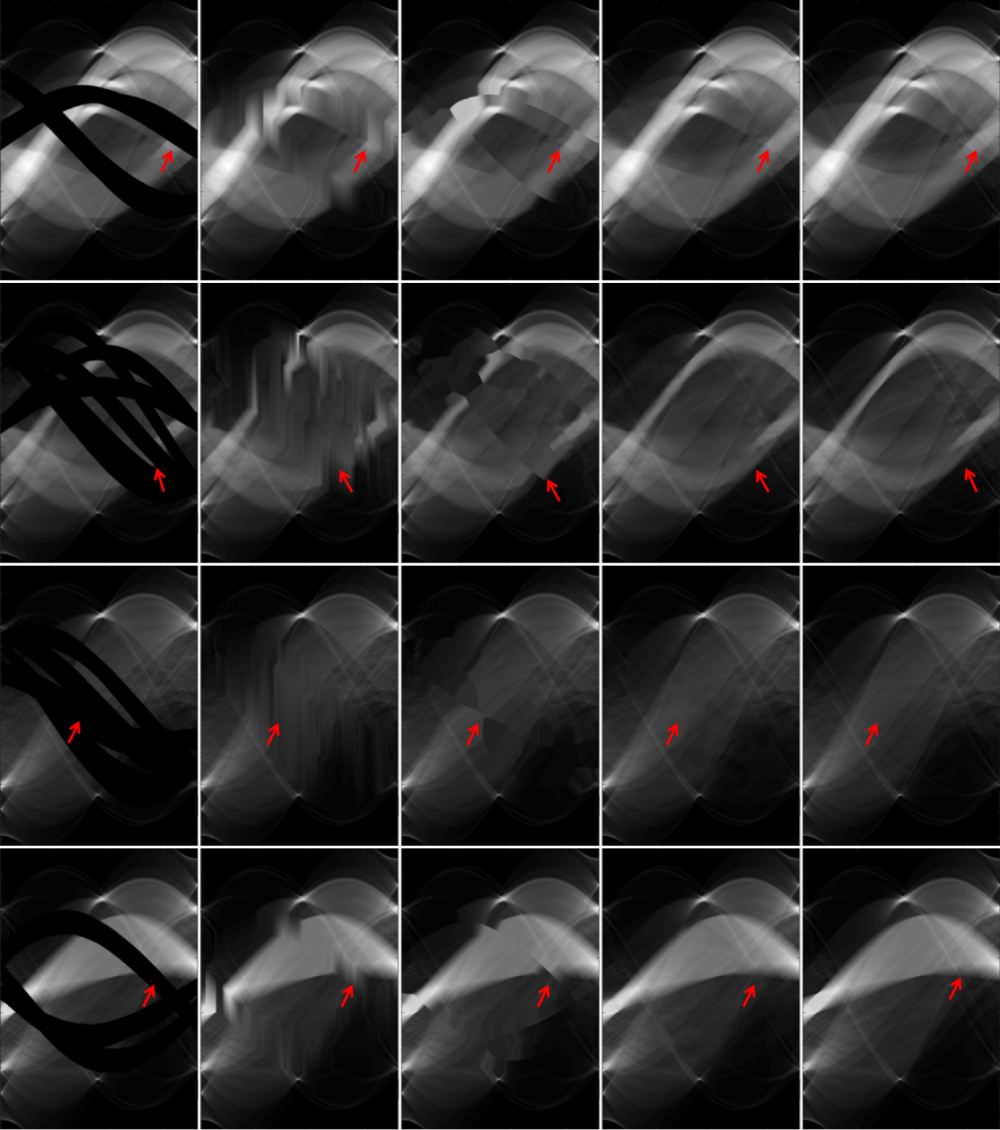}
\end{center}
\footnotesize
\hspace{5em}(a) Input\hspace{7.5em}(b) LI-MAR \cite{kalender1987reduction} \hspace{6em}WNN-MAR \cite{mahnken2003new}\hspace{5.25em}(d) \textbf{\emph{Deep}-MAR (Ours)} \hspace{2em} (e) Reference (metal-free)
\caption{Sinogram completion results for four real data examples from the ALERT Task Order $4$ \cite{crawford2014advances} are presented on display window $[0, 5]$. Incomplete sinograms which are used as input to sinogram completion methods are presented in column (a), sinogram completion results using LI-MAR, WNN-MAR, and our \emph{Deep}-MAR method are presented in columns (b), (c), and (d) respectively. Reference sinograms corresponding to metal-free scenes are presented in column (e). Reconstructions corresponding to completed and metal-free sinograms are presented in Figure \ref{fig:s_realRec_q}. These examples present a variety in number and geometric configuration of embedded metals and resulting holes in sinograms. While LI-MAR and WNN-MAR produce completed sinograms which are visibly bad and not consistent with reference sinograms, our \emph{Deep}-MAR performs excellent sinogram completion results which are consistent with reference sinograms.}
\label{fig:s_realSino_q}
\end{figure*}

\begin{figure*}[tb]
\begin{center}
\includegraphics[width = 0.99\textwidth]{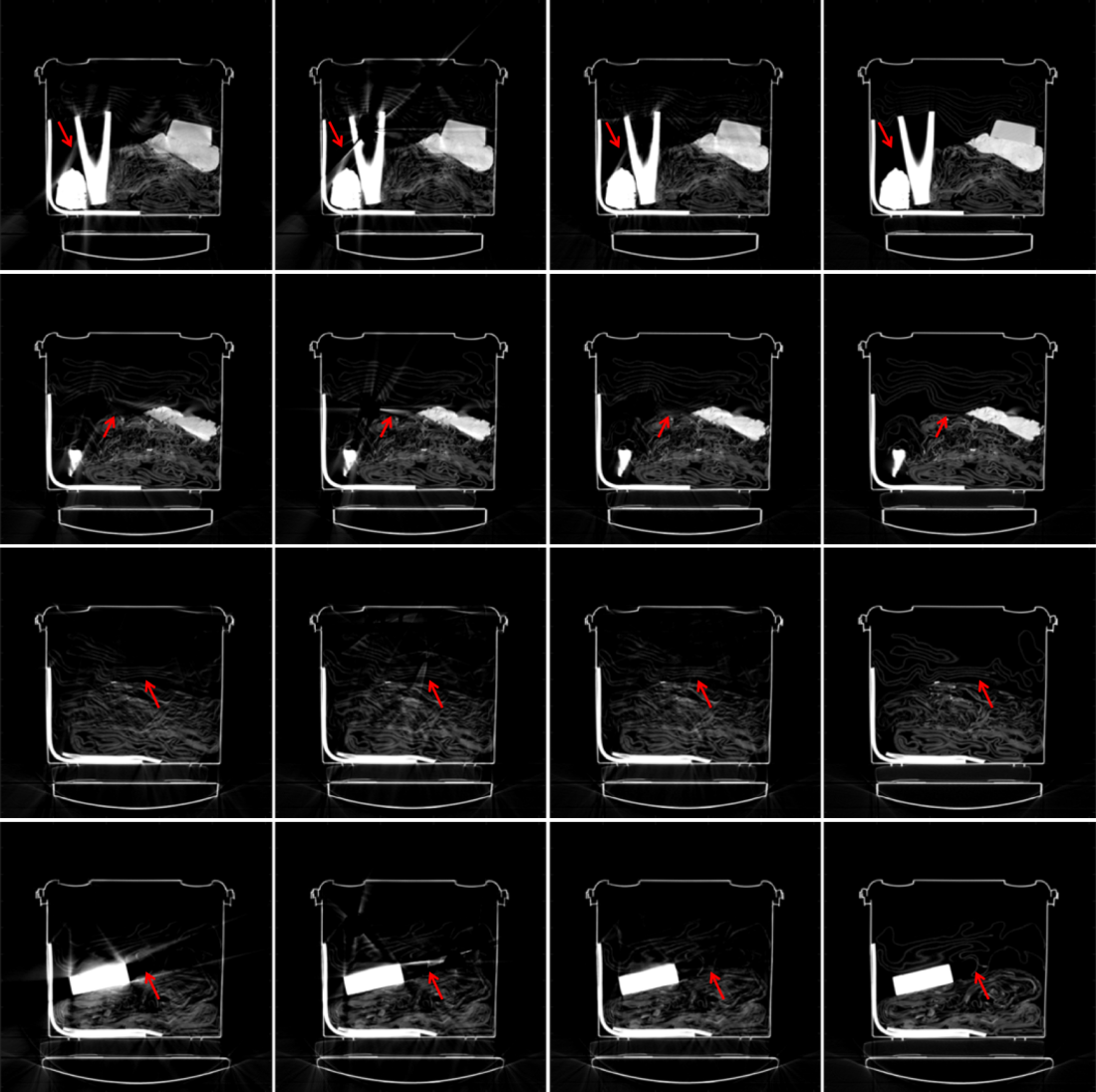}
\end{center}
\footnotesize
\hspace{5em}(a) LI-MAR \cite{kalender1987reduction} \hspace{8em}(b) WNN-MAR \cite{mahnken2003new} \hspace{8em}(c) \textbf{\emph{Deep}-MAR (Ours)} \hspace{4.5em}  (d) Reference (metal-free)
\caption{Reconstruction results for four real examples considered in Figure \ref{fig:s_realSino_q} are presented on display window $[0, 0.15] ~\text{ cm}^{-1}$. FBP reconstructions of sinograms completed with LI-MAR, WNN-MAR, and our \emph{Deep}-MAR are presented in columns (a), (b), and (c). FBP reconstructions of reference metal-free sinograms are presented in column (d). Example slices represent a variety in number, and type of materials, and varied level of clutter in the field of view. \emph{Deep}-MAR can reliably recover structure and suppress artifacts in all different scenarios.}
\label{fig:s_realRec_q}
\end{figure*}

Four additional test examples from the ALERT Task Order $4$ \cite{crawford2014advances} real dataset are presented for sinogram completion results in Figure~\ref{fig:s_realSino_q}. These slices do not contain metals, and incomplete sinogram have been generated by deleting data in the sinogram that corresponds to geometric configurations of embedded metal objects. We considered examples with a variety in the number of embedded metals and geometric configurations. While LI-MAR \cite{kalender1987reduction} and WNN-MAR \cite{mahnken2003new} struggle to correctly estimate the missing data in sinograms, our \emph{Deep}-MAR algorithm produces more accurate estimates of the missing data. This is evident by comparison of the completed sinograms to the reference sinograms presented in column (e). Reconstructions corresponding to different sinogram completion results considered in Figure \ref{fig:s_realSino_q} are presented in Figure \ref{fig:s_realRec_q} and reinforce these conclusions.  \par

\begin{figure*}[tb]
\begin{center}
\includegraphics[width = 0.9\textwidth]{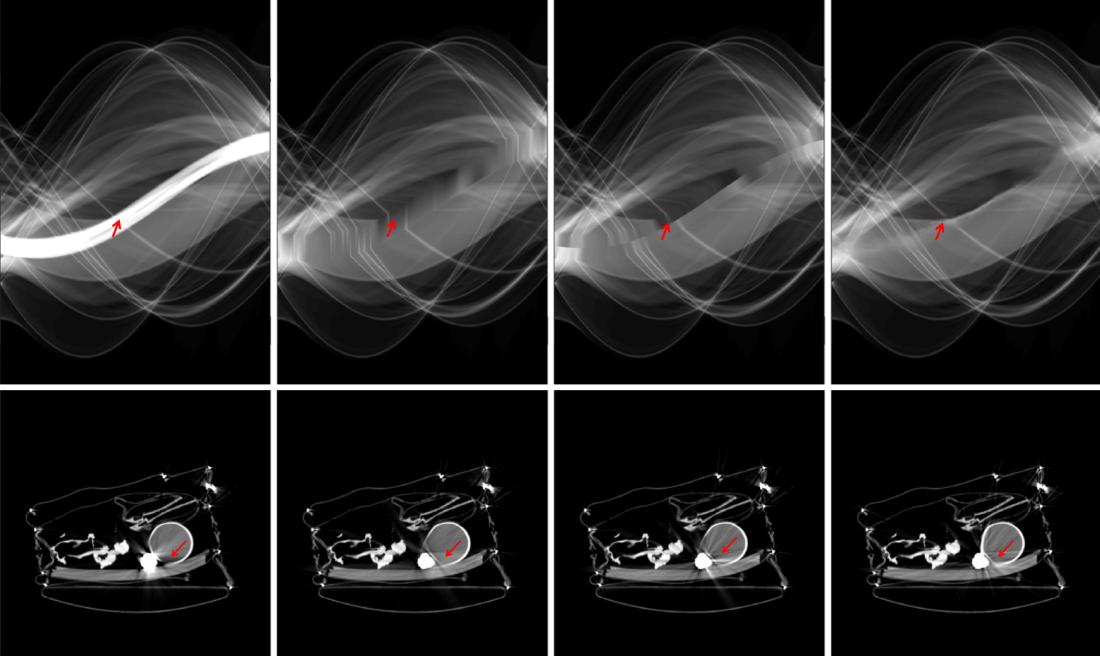}\\
\vspace{2pt}
\includegraphics[width = 0.9\textwidth]{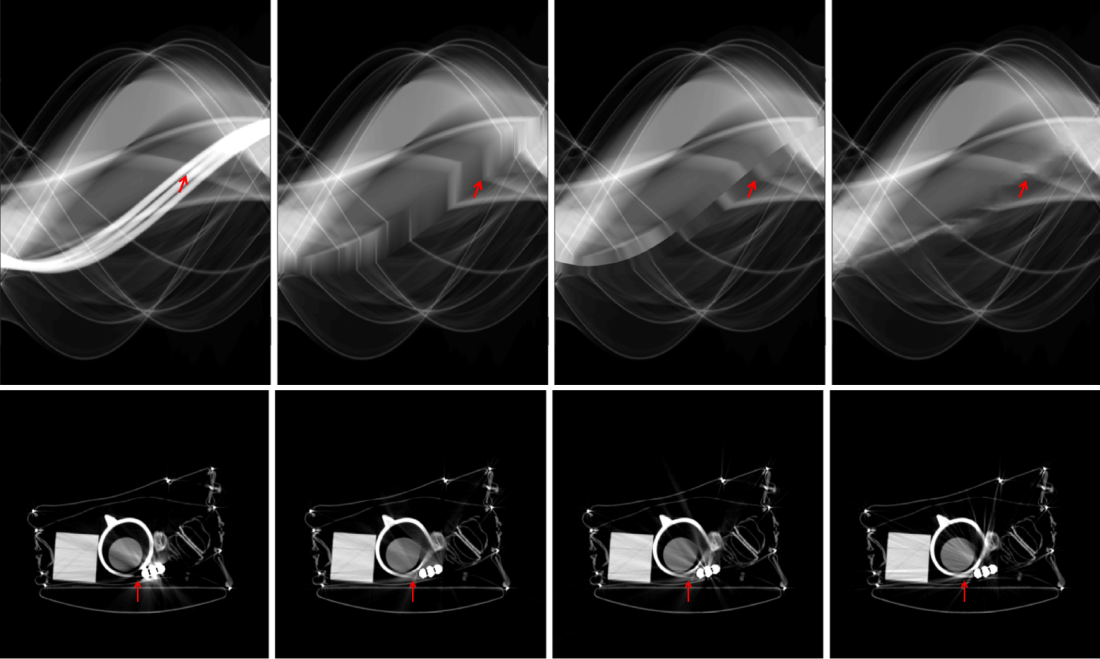}
\end{center}
\footnotesize
\hspace{6.5em}(a) Without MAR\hspace{8em}(b) LI-MAR \cite{kalender1987reduction} \hspace{8em}WNN-MAR \cite{mahnken2003new}\hspace{6.5em}(d) \textbf{\emph{Deep}-MAR (Ours)}
\caption{Sinogram completion and corresponding reconstruction results for two slices from the ALERT Task Order $3$ \cite{crawford2013research} real data with and without MAR techniques are presented on display window $[0, 7]$ and $[0, 0.4] ~\text{ cm}^{-1}$ respectively. Sinogram results are presented in first and third rows, and corresponding reconstructions are presented in second and fourth rows. Sinograms and corresponding FBP reconstructions without any MAR processing are presented in column (a). Sinograms and corresponding reconstructions without any MAR processing are presented in column (a), results for LI-MAR, WNN-MAR, and our \emph{Deep}-MAR are presented in column(b), (c), and (d) respectively. Red arrows mark areas which are of particular interest. It can be seen that metals produce dark shading and bright streaking artifacts in standard FBP reconstructions which result in object boundary suppression and non-uniformity in the object CT numbers. Both of these problems would effect the performance of automated object segmentation and identification algorithms. While LI-MAR and WNN-MAR suppress artifacts, the residual artifacts are still noticeable and they introduce new artifacts. Our \emph{Deep}-MAR can reliably recover structure lost in artifacts and greatly suppress metal artifacts.}
\label{fig:s_realRes1}
\end{figure*}

\begin{figure*}[tb]
\begin{center}
\includegraphics[width = 0.9\textwidth]{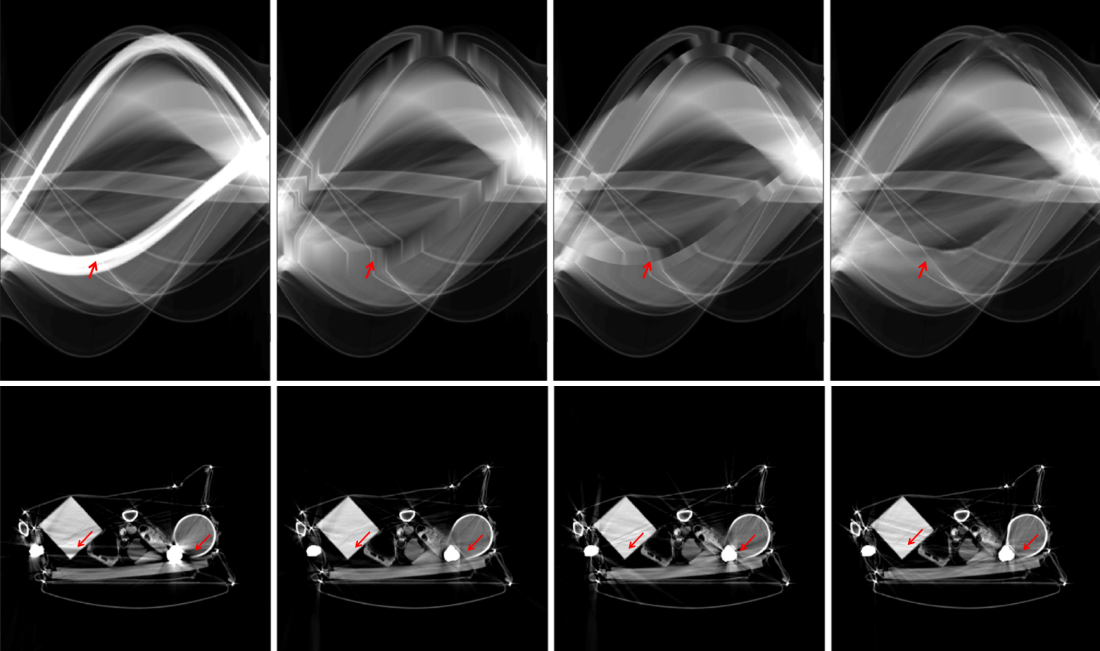}\\
\vspace{2pt}
\includegraphics[width = 0.9\textwidth]{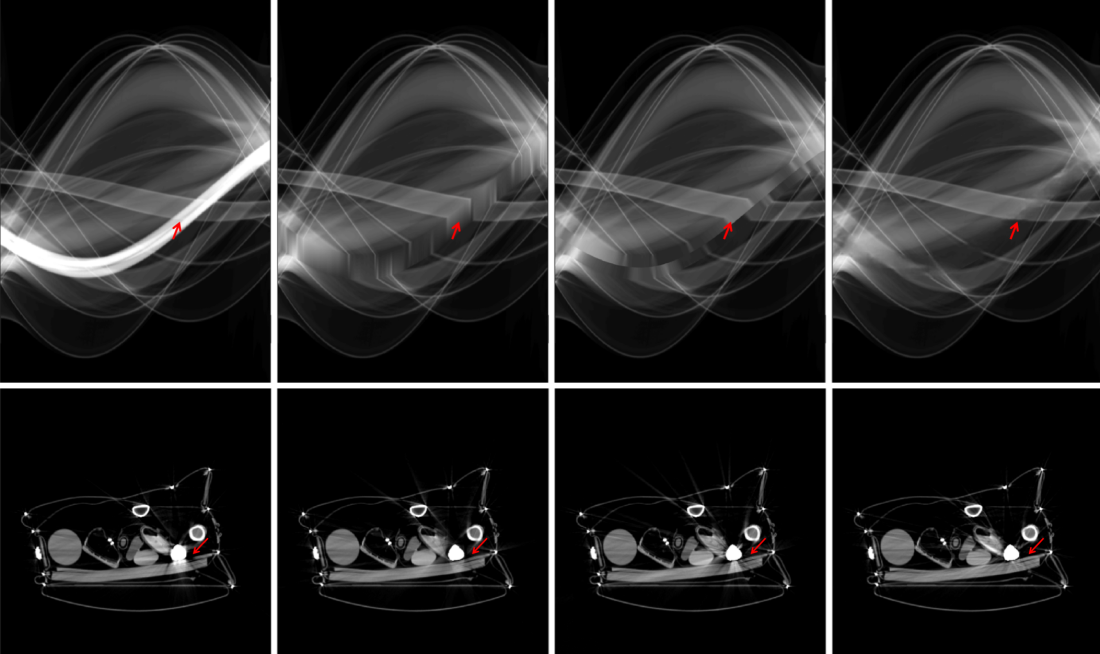}
\end{center}
\footnotesize
\hspace{6.5em}(a) Without MAR\hspace{8em}(b) LI-MAR \cite{kalender1987reduction} \hspace{8em}WNN-MAR \cite{mahnken2003new}\hspace{6.5em}(d) \textbf{\emph{Deep}-MAR (Ours)}
\caption{Sinogram completion and corresponding reconstruction results for two slices from he ALERT Task Order $3$ \cite{crawford2013research} real data with and without MAR techniques are presented on display window $[0, 7]$ and $[0, 0.4] ~\text{ cm}^{-1}$ respectively. Sinogram results are presented in first and third rows, and corresponding reconstructions are presented in second and fourth rows. Sinograms and corresponding FBP reconstructions without any MAR processing are presented in column (a). Sinograms and corresponding reconstructions without any MAR processing are presented in column (a), results for LI-MAR, WNN-MAR, and our \emph{Deep}-MAR are presented in column(b), (c), and (d) respectively. Red arrows mark areas which are of particular interest. It can be seen that metals produce dark shading and bright streaking artifacts in standard FBP reconstructions which result in object boundary suppression and non-uniformity in the object CT numbers. Both of these problems would effect the performance of automated object segmentation and identification algorithms. While LI-MAR and WNN-MAR suppress artifacts, the residual artifacts are still noticeable and they introduce new artifacts. Our \emph{Deep}-MAR can reliably recover structure lost in artifacts and greatly suppress metal artifacts.}
\label{fig:s_realRes2}
\end{figure*}

Further, we consider four test examples from the ALERT Task Order $3$ \cite{crawford2013research} real dataset and present sinogram completion and corresponding reconstructions with and without MAR methods in Figures \ref{fig:s_realRes1} and \ref{fig:s_realRes2}. The first and third rows present sinogram completions results, while the second and fourth rows present the corresponding reconstruction results. These example slices reflect a variety in number and type of materials and varied levels of clutter in the field of view. Our \emph{Deep}-MAR does a excellent job at the sinogram completion task. Since LI-MAR and WNN-MAR fail to correctly estimate sinogram completions, their corresponding reconstructions have residual artifacts, fail to completely recover the structure lost in streaks, and occasionally introduce new artifacts. On the other hand, our \emph{Deep}-MAR greatly suppresses artifacts and successfully recovers structure lost in streaks. It also preserves object boundaries and homogeneity inside the objects. \par

\section{Materials used for simulation} \label{supp:materials}

List of metallic and non-metallic materials along with chemical composition formulas used in generating simulated data are listed in this section. \par
\vspace{10pt}
\begin{tabular}{ll}
    % \hline
\textbf{Non-Metals }& \textbf{Metals} \\
Water ($H_2O$) & Silver ($Ag$)\\
Rubber ($C_5H_8$) & Copper ($Cu$)\\
Silicon ($Si$) & Tungsten ($W$)\\
Graphite ($C$) &Iron ($Fe$)\\
Teflon ($C_2F_4$) &Lead ($Pb$)\\
Delrin ($CH_2O$) &Tin ($Sn$)\\
Plastic ($C_2H_4$) &Mercury ($Hg$)\\
Acrylic ($C_5O_2H_8$) &Zinc ($Zn$)\\
PVC ($C_2H_3Cl$) & \\
Neoprene($C_4H_5Cl$) & \\
    % \hline
    \end{tabular}%
    
%   \subsection{Metals}
% \begin{itemize}
% \item Silver ($Ag$)
% \item Copper ($Cu$)
% \item Tungsten ($W$)
% \item Iron ($Fe$)
% \item Lead ($Pb$)
% \item Tin ($Sn$)
% \item Mercury ($Hg$)
% \item Zinc ($Zn$)
% \end{itemize}

% \subsection{Non-Metals} 
% \begin{itemize}
% \item Water ($H_2O$)
% \item Rubber ($C_5H_8$)
% \item Silicon ($Si$)
% \item Graphite ($C$)
% \item Teflon ($C_2F_4$)
% \item Delrin ($CH_2O$)
% \item Plastic ($C_2H_4$)
% \item Acrylic ($C_5O_2H_8$)
% \item PVC ($C_2H_3Cl$)
% \item Neoprene($C_4H_5Cl$)
% \end{itemize}

\small
\bibliographystyle{IEEEtran}
\bibliography{literature.bib}